\documentclass[10pt,journal,letterpaper,compsoc]{IEEEtran}
\ifCLASSOPTIONcompsoc
\else
\fi
\ifCLASSINFOpdf
\else
\fi
\let\labelindent\relax

\usepackage{times}
\usepackage{epsfig}
\usepackage{graphicx}
\usepackage{amsmath}
\usepackage{amssymb}
\usepackage{graphicx}
\usepackage{comment}
\usepackage{amsmath,amssymb} 
\usepackage[usenames,dvipsnames]{color}
\usepackage{enumitem}
\usepackage{booktabs}
\usepackage{colortbl}
\usepackage{tikz}
\usepackage{url}
\usepackage{ragged2e}
\usepackage[pagebackref=false,breaklinks=true,colorlinks,bookmarks=false]{hyperref}

\usepackage{algorithm}
\usepackage{algorithmicx}
\usepackage{algpseudocode}

\newcommand{\red}[1]{\textcolor{red}{#1}}
\newcommand{\blue}[1]{\textcolor{blue}{#1}}

\begin{document}
%
\title{Adaptive Siamese Tracking with a Compact Latent Network}
%
%
%
%

\author{Xingping~Dong, Jianbing~Shen,~\IEEEmembership{Senior Member,~IEEE},
Fatih~Porikli,~\IEEEmembership{Fellow,~IEEE},\\
Jiebo Luo,~\IEEEmembership{Fellow,~IEEE}, and Ling Shao,~\IEEEmembership{Fellow,~IEEE}

\IEEEcompsocitemizethanks{
\IEEEcompsocthanksitem This work was supported in part by the FDCT grant SKL-IOTSC(UM)-2021-2023,
the National Natural Science Foundation of China (No. 61929104),
the Start-up Research Grant (SRG) of University of Macau (SRG2022-00023-IOTSC),
and HPC resources in IIAI.
\IEEEcompsocthanksitem X. Dong is with the Inception Institute of Artificial Intelligence, Abu Dhabi, UAE.
(email: xingping.dong@gmail.com)
\IEEEcompsocthanksitem J. Shen is with the State Key Laboratory of Internet of Things for Smart City,
Department of Computer and Information Science, University of Macau, Macau, China.
(email: shenjianbingcg@gmail.com)
\IEEEcompsocthanksitem F. Porikli is with the Research School of Engineering, the Australian National University.
(email: fatih.porikli@anu.edu.au)
\IEEEcompsocthanksitem J. Luo is with the Department of Computer Scienece, University of Rochester.
(email: jiebo.luo@gmail.com)
\IEEEcompsocthanksitem L. Shao is with Terminus AI Lab, Terminus Group, China. (email: ling.shao@ieee.org)
\IEEEcompsocthanksitem A preliminary version of this work has appeared in ECCV 2020~\cite{dong2020clnet}.
\IEEEcompsocthanksitem Corresponding author: \textit{Jianbing Shen}
}
\thanks{}}

%

\markboth{IEEE Trans. Pattern Anal. Mach. Intell.} 
{Shell \MakeLowercase{\textit{et al.}}: Bare Demo of IEEEtran.cls for Computer Society Journals}
%


\IEEEcompsoctitleabstractindextext{%
\begin{abstract}
\justifying
In this paper, we provide an intuitive viewing to simplify the Siamese-based trackers by converting the tracking task to a classification. Under this viewing, we perform an in-depth analysis for them through visual simulations and real tracking examples, {and find that the failure cases in some challenging situations can be regarded as the issue of {\it missing decisive samples} in offline training. Since the samples in the initial (first) frame contain rich sequence-specific information, we can regard them as the decisive samples to represent the whole sequence. To quickly adapt the base model to new scenes, a compact latent network is presented via fully using these decisive samples. Specifically, we present a statistics-based compact latent feature for fast adjustment by efficiently extracting the sequence-specific information.} Furthermore, a new diverse sample mining strategy is designed for training to further improve the discrimination ability of the proposed compact latent network. Finally, a conditional updating strategy is proposed to efficiently update the basic models to handle scene variation during the tracking phase. To evaluate the generalization ability and effectiveness and of our method, we apply it to adjust three classical Siamese-based trackers, namely SiamRPN++, SiamFC, and SiamBAN. Extensive experimental results on six recent datasets demonstrate that all three adjusted trackers obtain the superior performance in terms of the accuracy, while having high running speed.
\end{abstract}

}

\maketitle

\IEEEdisplaynotcompsoctitleabstractindextext

%
\IEEEpeerreviewmaketitle

\section{Introduction}
Visual object tracking is a critical task in computer vision, with many applications such as automated surveillance, human-computer interaction, and vehicle monitoring~\cite{yilmaz2006object}. The objective of this task is to track a target, specified in the first few frames, throughout a video sequence. As with many other tasks of computer vision~\cite{krizhevsky2012imagenet,ren2015faster,he2017mask}, recent years have witnessed the huge success of deep learning methods in object tracking.

In the early years, researchers in the field focused on incorporating online correlation-filter-based models with deep features pre-trained on recognition tasks~\cite{krizhevsky2012imagenet}.
With the increase in labeled video data, more and more works~\cite{bertinetto2016fully-convolutional,li2018high,zhu2018distractor-aware} have attempted to re-train deep models on large-scale video datasets to obtain tracking-specific features, with SiamFC~\cite{bertinetto2016fully-convolutional} being one of the pioneering models. This work first introduced the large-scale video object detection dataset ILSVRC2015~\cite{russakovsky2015imagenet}, and then successfully incorporated data-driven deep learning with real-time visual tracking, attracting significant attention in the tracking community.
Several works have since followed, including  \cite{valmadre2017end--end,li2018high,zhu2018distractor-aware,fan2019siamese,li2019siamrpn,zhang2019deeper}. For instance, Li~{\it et al.}~\cite{li2018high} introduced the Region Proposal Network (RPN) into the Siamese framework, avoiding the requirement of multiple computations for the scale estimation of SiamFC, and achieving faster speed (160 frame-per-second (FPS)) and better tracking performance. Many variants based on this have also been proposed, such as SiamRPN++~\cite{li2019siamrpn}, which further enhances the accuracy while maintaining the high speed.

\begin{figure}[!htp]
	\centering
	\includegraphics[width = .492 \textwidth]{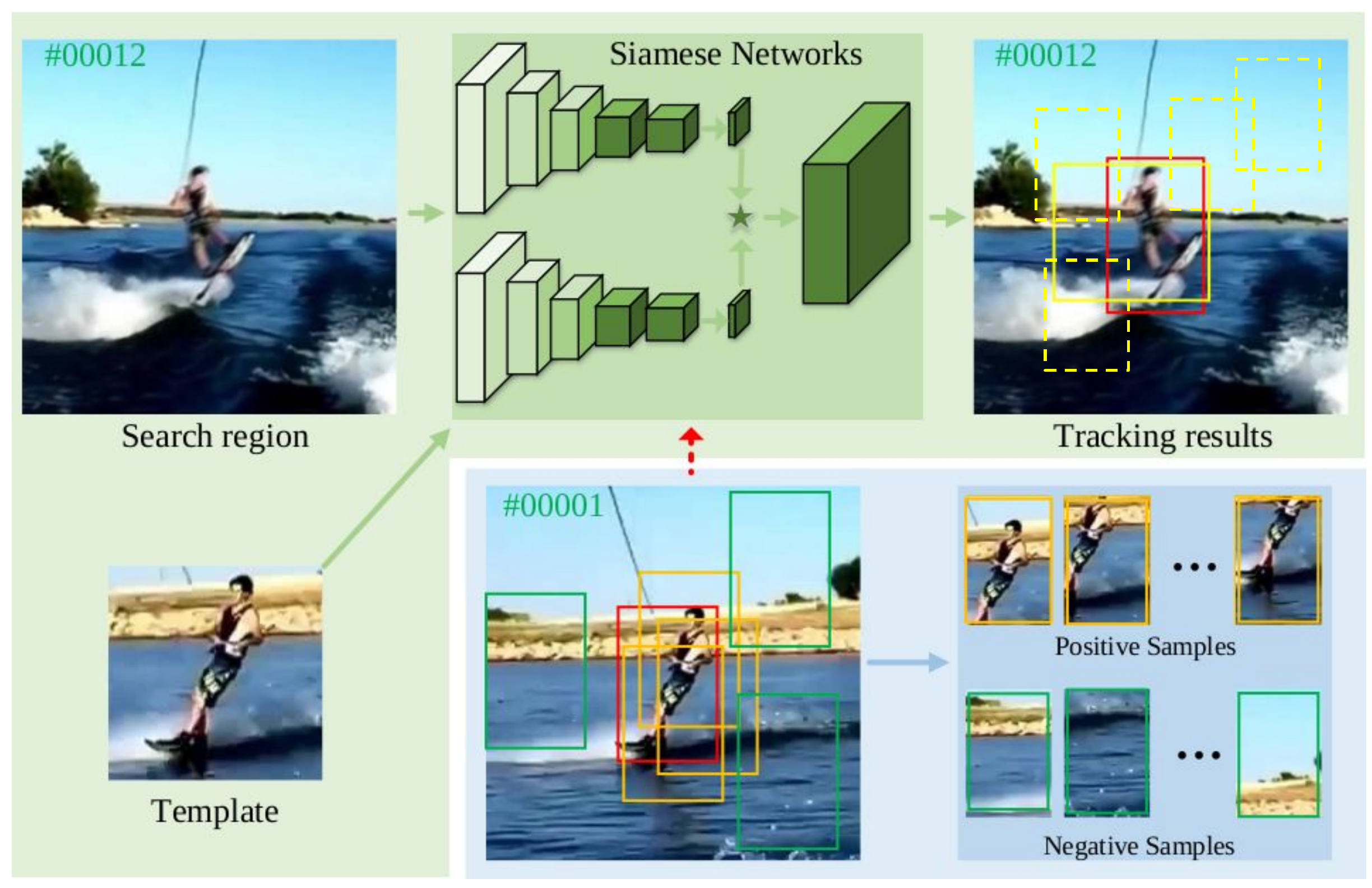}\\
	\caption{{\small\textbf{Illustration of tracking procedure (green panel) and ignoring context (blue panel) of a Siamese-based model (SiamRPN++~\cite{li2019siamrpn}).} In the green panel, the red solid box is the ground truth, the yellow one is the final predicted bounding box (bbox) and the yellow dotted boxes are candidate samples. The blue panel shows the sequence-specific samples, which are extracted from the first frame. Green and orange rectangles indicate the positive and negative bboxes, respectively. However, most Siamese-based approaches ignore these sequence-specific samples in the first frame, leading to failures in challenging cases, such as frame {\#}0012 in this example.}
	}
	\label{fig:motivation}
\end{figure}

However, these Siamese-based trackers are still unable to address more challenging situations, such as the presence of similar distractors or significant deformation.
We provide a new and intuitive method to analyze the underlying reason for these failures. Specifically, we simplify the Siamese-based model to a linear binary-classifier and convert the tracking task to a classification problem. For example, as shown in the tracking results in Fig.~\ref{fig:motivation}, a Siamese-based model (such as SiamRPN++~\cite{li2019siamrpn}) usually produces many boxes (yellow dotted boxes) with classification scores as candidates and then selects the box with the highest score as the final tracking result (yellow solid box). We can regard the image patches under these boxes as candidate classification samples. Then, the tracking problem is transferred to classifying these candidate samples.
{In this case, we observe that above failures come from the issue of {\it missing decisive samples}, \textit{i.e.} some key samples, such as the samples in these challenging cases, are rare or unseen during offline training.
This will lead to the poor discriminative ability of the trained model for these samples. This is a common problem with most data-driven models, since they do not capture all the data used for training. {In contrast, in the tracking task, each sequence provides annotated bounding boxes (bboxes) in the first frame. These boxes can generate sequence-specific samples and improve the recognition ability of the model, since these samples include most foreground and background information in this sequence.} For example, in Fig.~\ref{fig:motivation}, we randomly draw several bboxes in the first frame to produce the sequence-specific samples, which are labeled as positive or negative ones according to their overlap with the ground-truth. These samples, such as `sky' or `sea' patches in our example, can provide important context in other frames (frame {\#}0012) to facilitate the tracking.
However, most Siameses-based methods only apply annotations to extract templates, thus ignoring the rich contextual information. {Therefore, these models can be tuned with neglected contextual information, improving their discriminative ability for all samples in the sequence}.
}

{However, fully utilizing sequence-specific information to tune a model trained offline is a challenging task, especially for real-time tracking, since it incurs computational load. A simple solution is to use an optimization method such as stochastic gradient descent (SGD) ~\cite{nam2016learning}, ride regression ~\cite{danelljan2016beyond}, or Lagrangian multipliers ~\cite{hong2015online}, directly retrain the Siamese-based model. However, these methods are usually impractical and time-consuming for tracking.}

{To extract effective information from sequence-specific samples and meet real-time requirements, a compact latent network (CLNet) is proposed to tune recent Siamese-based trackers. The proposed CLNet consists of a feature-adjusting subnetwork, a latent encoder, and a prediction subnetwork. The first module is built by three $1 \times 1$ convolutional layers for efficiency, and provides the adjusted feature for each sequence-specific sample. The core module, the latent encoder, generates a compact feature representation for the entire set of sequence-specific samples by computing the statistical information inside the positive and negative adjusting feature sets, respectively. Next, the latent features are fed into the last subnetwork (a three-layer perceptron) to predict the adjusting parameters.}

{It is worth mentioning that our statistics-based latent features are compact and only with a few thousand parameters. Furthermore, they include more uncertainty information than standard features, which is beneficial for performance improvement~\cite{khan2019striking}. In addition, a new diverse sample mining algorithm is proposed for training to further enhance the performance. In contrast to the previous training methods~\cite{li2019siamrpn}, we aim to effectively capture the sequence-specific information to train the latent feature. Thus, for each batch, we adopt image-pairs sampled from a sequence.
To handle significant appearance changes and similar distractors, we mine the diverse samples to enhance the training performance.}

{To verify the effectiveness, we adopt three representative trackers, SiamRPN++~\cite{li2019siamrpn}, SiamFC~\cite{bertinetto2016fully-convolutional}, and SiamBAN~\cite{chen2020siamese}, as our baselines.
To keep the generalization ability of the baseline models, our CLNet is applied to adjust the last layers in the regression and classification branches. For handling scene variation and fast adjustment, we propose a new conditional updating strategy to dynamically adjust basic models, by using classification scores of candidate boxes.
Specifically, CLNet is only used on the first frame and the frames that suffer from scene changes, as determined by our conditional update strategy, and is ignored in other frames.
Comprehensive evaluations on six popular benchmarks (in \S\ref{sec:experiment}) demonstrate the strengths of the proposed algorithm.
}

Our key contributions are as follows:
\begin{itemize}
	\item We provide an {\bf in-depth analysis of Siamese-based trackers} and demonstrate that their poor discrimination ability under certain challenging scenarios is caused by {\it  missing decisive samples} during training. We find that this problem can be alleviated by utilizing sequence-specific samples from the first frame. 
	\item A {\bf novel framework termed CLNet} is proposed to adjust deep Siamese-based trackers by efficiently capturing the sequence-specific information. To our knowledge, this is the first attempt to utilize statistics-based latent features to adjust deep trackers.
	\item We incorporate a {\bf diverse sample mining} technique to facilitate network training. This technique can effectively enhance the discrimination ability of CLNet, by capturing the critical statistical information of a sequence.
	\item { A \textbf{conditional updating strategy} is designed for fast and robust adjustment. This strategy can dynamically adjust basic models to efficiently utilize sequence-specific information and handle scene variation.
	}
	\item
	{Comprehensive experiments on six tracking benchmarks show the benefits of our CLNet, which leads to {\bf promising improvement for three baseline trackers}: SiamRPN++, SiamFC, and SiamBAN.
	Moreover, our adjusted trackers can still remain high running speed over 38 FPS.}
	
\end{itemize}

This paper significantly extends our previous work in \cite{dong2020clnet},
and presents several improvements with extensive discussions.
{First, we obtain additional details of our approach via an intuitive illustration of the missing decisive samples and ignoring context caused by current Siamese-based models to better illustrate our motivation.
Second, we provide thorough discussions analyzing the core causes for the failures in Siamese-based models by conducting qualitative and quantitative analyses on real sequences.
Third, we design a new conditional updating strategy to dynamically and efficiently adjust basic models for further performance improvement and remaining high speed.
Fourth, we apply our method to two other representative Siamese-based trackers, SiamFC and SiamBAN, with further elaboration on the formulations and the implementation details.
Fifth, we provide more experiments for ablation study to analyze the impact of adjusting regression branches and different weight augmentation approaches.
Last but not least, we add two new datasets, VOT2020~\cite{Kristan2020a} and GOT10k~\cite{huang2019got-10k} and conduct extensive experiments to thoroughly examine the effectiveness in quickly adjusting Siamese-based trackers. The results on SiamRPN++, SiamFC, and SiamBAN demonstrate the generalization ability of our adjustment method.}

\section{Related Work}\label{sec:relate work}
{Although many methods have been proposed for visual object tracking, such as the regression model~\cite{held2016learning,lu2018deep}, correlation filter~\cite{henriques2015high-speed,danelljan2017eco,dong2017occlusion-aware}, and sparse coding~\cite{ma2015visual,ma2015linearization}.
Here, we pay attention to Siamese networks based trackers~\cite{bertinetto2016fully-convolutional,li2018high}, and meta-learning~\cite{hinton1987using,schmidhuber1987evolutionary}, mostly related to our algorithm.}

\subsection{Siamese Networks Based Trackers}
In early years, pre-trained CNN models were used to extract powerful features and incorporated with different online tracking methods~\cite{wang2015visual,ma2015hierarchical,danelljan2016beyond,tao2016siamese,danelljan2017eco,shen2019visual}.
As the increment of the available datasets for tracking~\cite{russakovsky2015imagenet,lin2014microsoft,real2017youtube-boundingboxes,fan2019lasot,huang2019got-10k}, more and more researchers have paid attention to designing new networks and retraining their models with tracking datasets~\cite{nam2016learning,held2016learning,bertinetto2016fully-convolutional,zhang2018synthetic,li2018high,li2019siamrpn}. Among these approaches, trackers based on Siamese networks~\cite{tao2016siamese,bertinetto2016fully-convolutional} have become mainstream in recent years. 

{Bertinetto \textit{et al.}~\cite{bertinetto2016fully-convolutional}, provided a new paradigm to train fully convolutional Siamese networks (SiamFC), by using the large-scale labeled video datasets. Specifically, SiamFC learns a similarity metric based on the Siamese networks, through offline training on a large-scale video detection dataset~\cite{russakovsky2015imagenet}. Then, the learned Siamese networks are directly applied to generate the score map between the target and search region for tracking, without any online training. This paradigm, separating offline training and online tracking, achieves a well balance between efficiency and accuracy.
{Since this work, more attentions are attracted to further mine the offline tracking model by presenting various Siamese networks~\cite{he2018twofold,wang2018learning,zhang2018structured,dong2019quadruplet}, using efficient Siamese networks~\cite{shen2021distilled}, designing a powerful training loss~\cite{dong2018triplet}, and so on~\cite{wang2018sint,yang2018learning,shen2019visual}. Some researchers have focused on presenting various strategies for online updating, such as utilizing deep reinforcement learning~\cite{huang2017learning,dong2018hyperparameter,dong2019dynamical} , learning a dynamic network~\cite{guo2017learning}, or incorporating a correlation filter~\cite{valmadre2017end--end}.}
Among these trackers, SiamRPN~\cite{li2018high} successfully applies the region proposal network to Siamese networks, achieving impressive accuracy on the challenging VOT dataset~\cite{kristan2016novel}, and with particularly high speed. Several following works also set up the new state-of-the-art on it~\cite{zhu2018distractor-aware,fan2019siamese,li2019siamrpn,zhang2019deeper}. For instance, Zhu~{\it et al.}~\cite{zhu2018distractor-aware} presented an incremental learning approach and a distractor-aware data augmentation method to enhance the online tracking mechanism and offline training, respectively. Since this, more researchers focus on improving the discriminative ability of the core network by designing more powerful architectures, such as a deeper networks~\cite{li2019siamrpn}, cascaded RPN~\cite{fan2019siamese}, and wider networks~\cite{zhang2019deeper}. These methods prefer to propose more discriminative offline models, utilizing more data. However, the contextual information of the first frame is ignored.
}

\begin{figure*}[ht]
	\centering
	\includegraphics[width = .9 \textwidth]{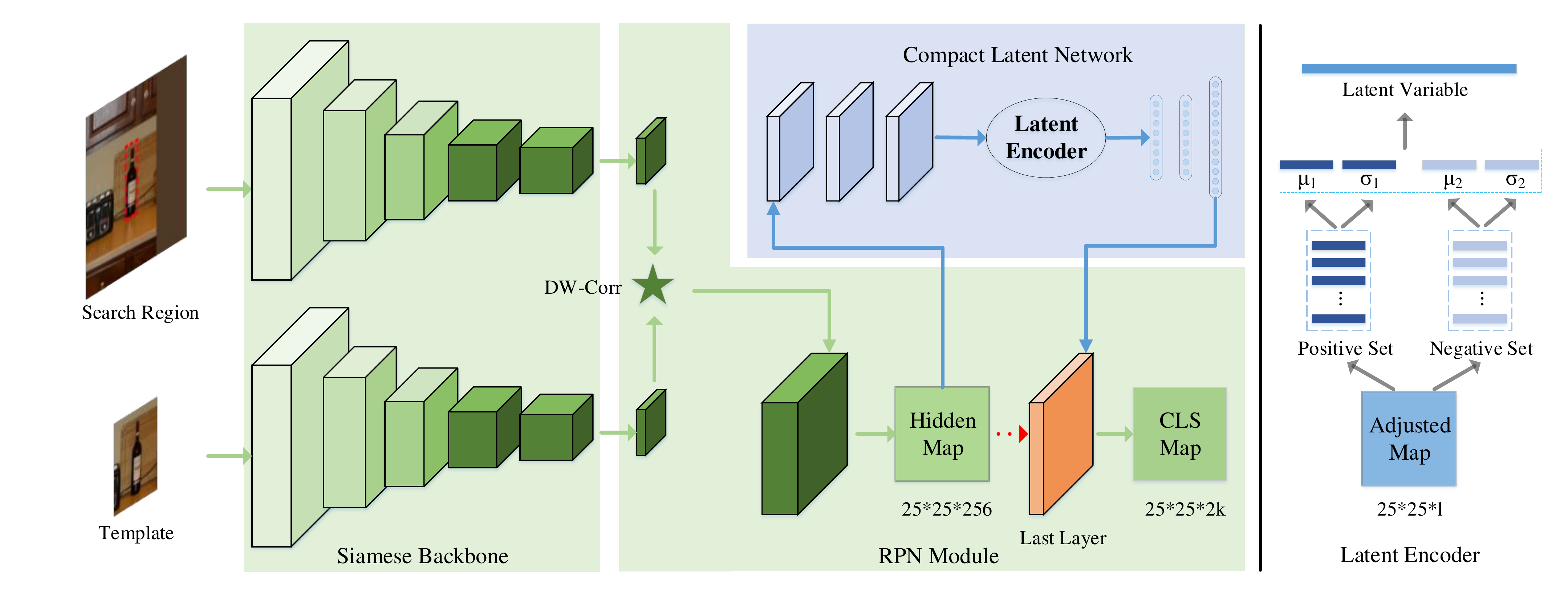}
	\caption{\small {{\bf Flowchart of adjusting a Siamese-based tracker (such as SiamRPN++~\cite{li2019siamrpn}) via our compact latent network (CLNet).} For clarity, only the classification branch is shown. In the first frame, our CLNet predicts the weights of the last layer to adjust the basic model by using the hidden map. In the following frames, we only adjust the basic model on a few frames by using our conditional updating strategy (\S\ref{sec:cu}) to reduce computation and handle scene variation. In most frames, we omit the CLNet and apply the adjusted model for tracking. We show the latent encoder on the right, where $\sigma$ and $\mu$ are the standard deviation and mean of a sample set. We generate the positive and negative sets using the annotated (predicted) bounding box in the first (previous) frame.}}
	\label{flowchart}
\end{figure*}

\subsection{Meta-Learning}
{
{\it Meta-learning} is usually interpolated as the inception of {\it learning-to-learn}~\cite{schmidhuber1987evolutionary,thrun1998learning,hochreiter2001learning,andrychowicz2016learning}, or {\it fast weights}~\cite{hinton1987using,ba2016using}, and can be applied to the few-shot recognition task. Recent meta-learning approaches are roughly grouped by three major categories, including 1) memory-based methods~\cite{santoro2016meta-learning,ravi2017optimization}, which investigate the storage of key training examples by encoding fast adaptation methods or utilizing effective memory architectures;  2) optimization-based methods~\cite{finn2017model-agnostic,finn2018probabilistic}, which explore the adaptive parameter initialization to quickly tune the model for new tasks; and 3) metric-based methods~\cite{koch2015siamese,vinyals2016matching,snell2017prototypical}, which focus on learning powerful similarity metrics to discriminate samples from the same class.
}

{Although recent years witness the success of meta-learning in few-shot classification~\cite{finn2017model-agnostic,rusu2019meta-learning,li2019lgm-net}, only a few works are related with visual object tracking~\cite{park2018meta-tracker,choi2019deep}. An optimization-based meta-learner~\cite{park2018meta-tracker} is proposed for gradient-based online updating by learning an adaptive step. This approach has been used to speedup two online training trackers~\cite{nam2016learning,song2018vital}, by decreasing the training iterations. However, it is not suitable for offline training models, such as Siamese-based trackers. Recently, a gradient-guided network~\cite{li2019gradnet} and meta-learner model~\cite{choi2019deep} were presented to online update Siamese networks based trackers, by capturing target-specific information with the gradients. In contrast, our algorithm investigates a new direction to use the statistics-based latent feature to extract the sequence-specific information. }

\section{Siamese-based Tracker with Compact Latent Network}\label{sec:siamese tracker}
We adopt SiamRPN++~\cite{li2019siamrpn} as our basic tracker, which is a recent representative Siamese-based tracker. Its framework is briefly introduced in \S\ref{sec:siamrpn}. Then, a deep analysis is conducted to explore the underlying issue of Siamese-based trackers, called as {\it missing decisive samples} (see \S\ref{sec:analysis}). To alleviate this problem, an efficient compact latent network (CLNet) is proposed in \S\ref{sec:latent_metanet}, by adjusting the basic model. Furthermore, a new mining method is proposed to find the diverse training samples in \S\ref{sec:hard sample}.
{We show our framework in Fig.~\ref{flowchart}.} Besides, we combine it to the more recent state-of-the-art SiamBAN~\cite{chen2020siamese} tracker, and classical SiamFC~\cite{bertinetto2016fully-convolutional} tracker in \S\ref{sec:extension}, to further verify the generalization ability of our approach.

\subsection{Revisiting SiamRPN++ for Tracking}\label{sec:siamrpn}
The pioneering SiamFC~\cite{bertinetto2016fully-convolutional} regards the target patch as a template, which is given in the first frame, and the goal is to find the most similar patch from the adjacent (search) region of the following frames. SiamFC~\cite{bertinetto2016fully-convolutional} construct a fully convolutional Siamese model $\phi$ to obtain representative features. This model includes an instance branch to process the search region $\mathbf{x}$, and a template branch to represent the target patch $\mathbf{z}$. Then the final similarity map $\mathbf{S}$ is generated by the cross-correlation (convolution) operation:
\begin{equation}\label{siamfc}
\mathbf{S}(\mathbf{x},\mathbf{z})=\phi (\mathbf{x}) * \phi(\mathbf{z})+b,
\end{equation}
where $*$ represents the cross-correlation operation and $b$ is the offset of the similarity value. $\mathbf{S}$\footnote{For simplification, we remove $(\mathbf{x},\mathbf{z})$ for similar math notations.
} $\in \mathbb{R}^{w \times h}$ is the similarity metric for the searching instances and template, where $w$ and $h$ represent the height and width of this map. Then the target location is on the peak of $\mathbf{S}$.

SimRPN~\cite{li2018high} incorporates Region Proposal Network (RPN) to extend SiamFC. It simultaneously makes position and scale estimation by regressing the target bounding box (bbox) with the classification and regression branches. It also uses a multi-anchors technique~\cite{ren2015faster} for higher accuracy. Incorporated with Siamese networks, SiamRPN generates $w \times h \times k$ anchors, where each search position will produce $k$ anchors on search feature map $\phi(\mathbf{x})$. The classification and regression branches provide a classification score and corresponding proposal (bbox) for each anchor.
The formulations 
are as follows:
\begin{equation}
\begin{aligned}
\mathbf{A}^{cls}_{w \times h \times 2k} & = conv_{cls}^{fea}(\phi(\mathbf{x})) * conv_{cls}^{ker}(\phi(\mathbf{z})),\\
\mathbf{A}^{loc}_{w \times h \times 4k} & = conv_{loc}^{fea}(\phi(\mathbf{x})) * conv_{loc}^{ker}(\phi(\mathbf{z})),
\end{aligned}
\end{equation}
{where $conv$ represents the convolutional network} providing new feature maps. 
Then the proposal with the highest classification score is the final target bbox. To reduce the number of parameters, SiamRPN++~\cite{li2019siamrpn} uses the asymmetrical depth-wise cross-correlation to provide efficient computation and also alleviates the issue caused by the distinction between the regression and classification. The formulation of new RPN block is as follows:
\begin{equation}
\begin{aligned}
\mathbf{A}^{cls}_{w \times h \times 2k} & = head^{cls}\left(\alpha_{cls}^{fea}(\phi(\mathbf{x})) \star \alpha_{cls}^{ker}(\phi(\mathbf{z}))\right), \\
\mathbf{A}^{loc}_{w \times h \times 4k} & = head^{loc}\left(\alpha_{loc}^{fea}(\phi(\mathbf{x})) \star \alpha_{loc}^{ker}(\phi(\mathbf{z}))\right),
\end{aligned}
\label{eq:rpn-map}
\end{equation}
where $\star$ is the depth-wise cross-correlation, $\alpha$ is an adjusted ($1 \!\times\! 1$ convolutional) layer, and $head$ denotes the head block to predict maps of regression and classification. 

\subsection{Analysis of Siamese-Based Training Methods} \label{sec:analysis}
Siamese-based trackers are trained by using numerous image pairs  from labeled video datasets (e.g., VID~\cite{russakovsky2015imagenet}) as rich training samples.
The generalization ability of tracking models can be effectively enhanced by these rich samples. However, the sequence-specific information is ignored, which is given by the annotation in the initial frame. Here, a simple binary-classification case is introduced to intuitively illustrate the underlying issue in the general training method. We can also regard SiamFC~\cite{bertinetto2016fully-convolutional} as a binary-classification model. Specifically, the template is viewed as the classifier to classify the instance patches. Thus, our analysis is suitable for most approaches based on SiamRPN~\cite{li2018high} or SiamFC~\cite{bertinetto2016fully-convolutional}.

\begin{figure}[!tbh]
	\centering
	\includegraphics[width = .45 \textwidth]{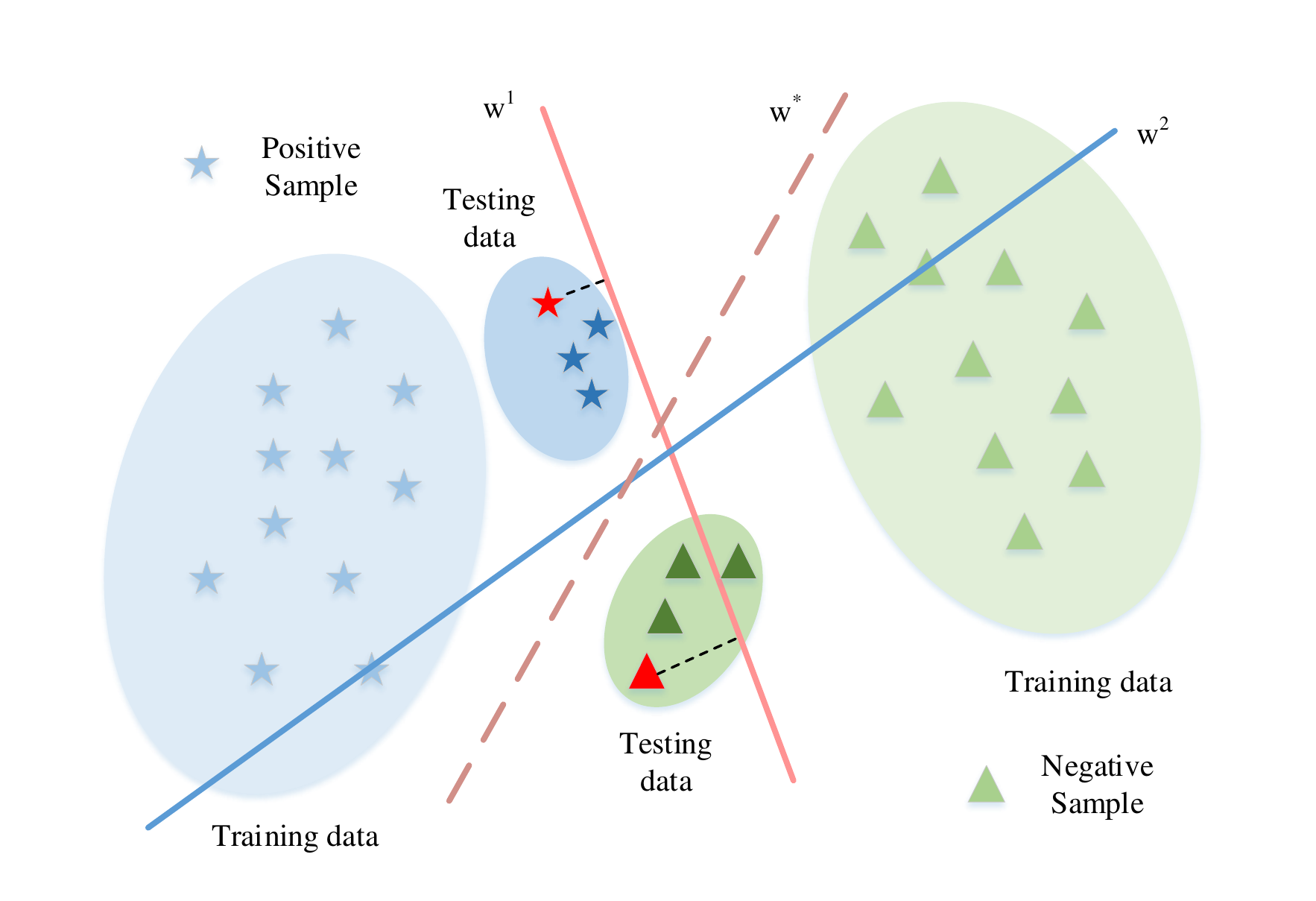}
	\caption{\small {\bf The binary-classification issue with the general Siamese-based training method.} $w^1$ and $w^2$ are the ideal decision hyperplanes based on the training and testing sets, respectively. The dotted $w^*$ is the decision hyperplane considering both sets. The red star and triangle represent the decisive samples.}
	\label{example}
\end{figure}

\noindent\textbf{Visualization Analysis.} According to the Siamese-based training method, we assume that various negative and positive samples come from different videos in training. As shown in Fig.~\ref{example}, a group of negative samples (the big green ellipse) and positive samples (the big blue ellipse) are used for training. To maintain the largest margin of the two group boundaries, the optimal decision hyperplane should be in the middle of the two groups. $w^1$ is assumed as the ideal decision hyperplane based on the training data in Fig.~\ref{example}. In testing, we assume the testing negative and positive samples (e.g. sampled from unseen videos) do not overlap training samples, and the testing samples are usually far fewer than training samples. Therefore, the testing samples are represented by small ellipses, and the challenging samples are assumed to lie near the decision hyperplane. Some challenging samples in Fig.~\ref{example} may pass through the decision hyperplane. This indicates the trained classifier can not discriminate against these difficult samples. 

{If any negative sample's score is higher than some positive samples in the general classification task, an error will occur. In contrast, in the tracking task, the error occurs only when one negative sample's score is higher than all positive samples.
However, there still} exists the problem of an unsuitable decision hyperplane (e.g. $w^1$) in tracking. For example, as shown in Fig.~\ref{example}, we take samples from the search region as the testing data points. Among all negative samples, we denote the point farthest away from the decision hyperplane $w^1$ as the red triangle. This indicates it has the highest positive score. Similarly, among all positive samples, the red star means the point with the highest score. We can find the red triangle's score is higher than that of the red star. Therefore, the red triangle is viewed as the final target, leading to tracking faults.

{A simple method to alleviate the above issue is using some optimization models to retrain the classifier with the available testing data (e.g. the support vector machine). However, this method may ignore the information captured from training data and reduce the classifier's generalization, because only a few samples are available for retraining. For example, in Fig.~\ref{example}, if only a few testing samples (\textit{i.e.}, the ones in the first frame) are used for retraining, the decision hyperplane should be $w^2$, not the ideal $w*$ (considering testing and training data). This means that retrained classifier fails to distinguish some difficult samples, like the triangles in the up-right of ellipse (training data), even if it has learned them during training. Thus, a new solution should be explored for this issue.}

\begin{figure}[!htp]
	\centering
	\includegraphics[width = .24 \textwidth]{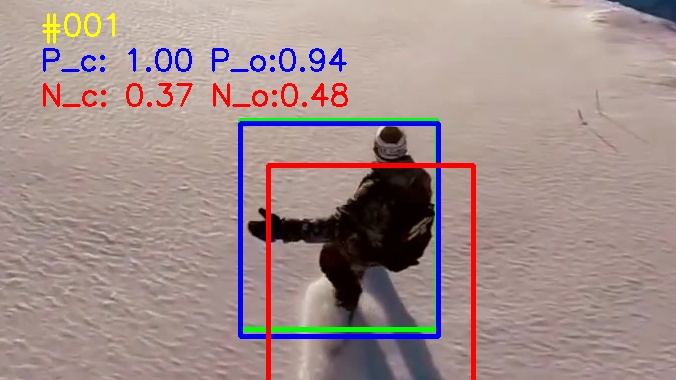}
	\includegraphics[width = .24 \textwidth]{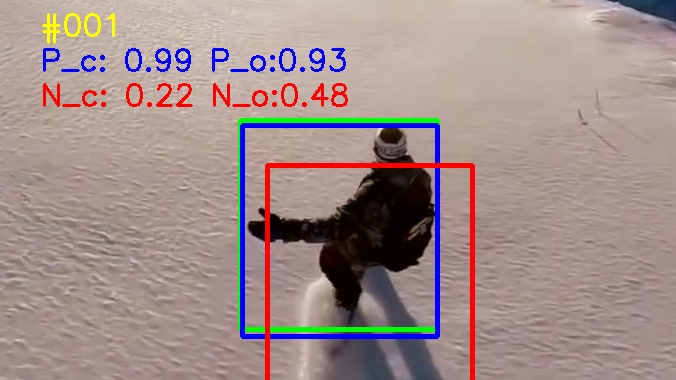}\\
	\includegraphics[width = .24 \textwidth]{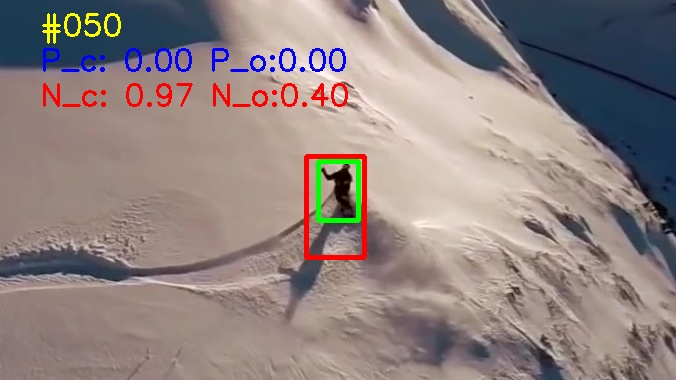}
	\includegraphics[width = .24 \textwidth]{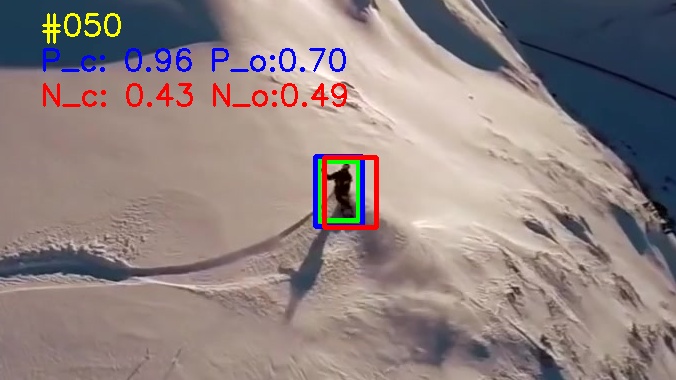}
	\\
	\scriptsize
	{~~(a) ~~~~~~~~~~~~~~~~~~~~~~~~~~~~~~~~~~~~~~~~~~~ (b)}\\
	\includegraphics[width = .24 \textwidth]{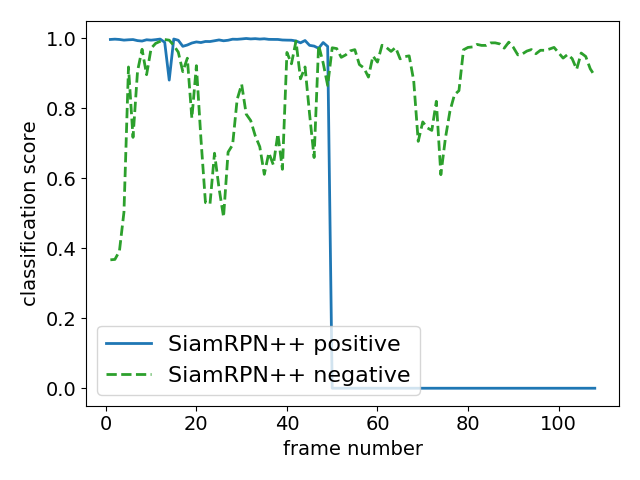}
	\includegraphics[width = .24 \textwidth]{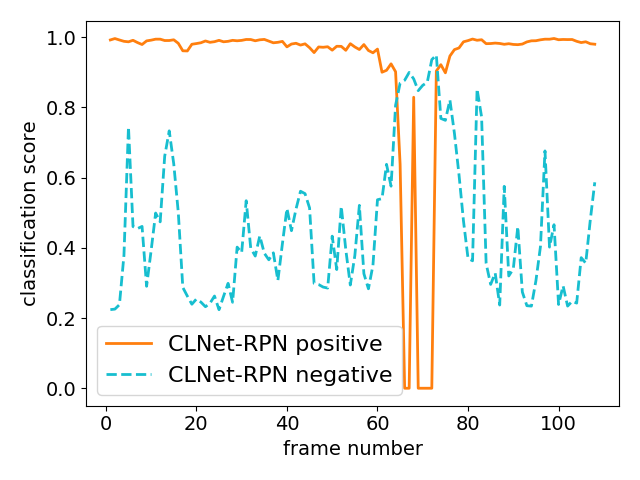}\\
	~~(c)~~~~~~~~~~~~~~~~~~~~~~~~~~~~~~~~~~~~~~~~~~~ (d)
	\\
	\includegraphics[width = .24 \textwidth]{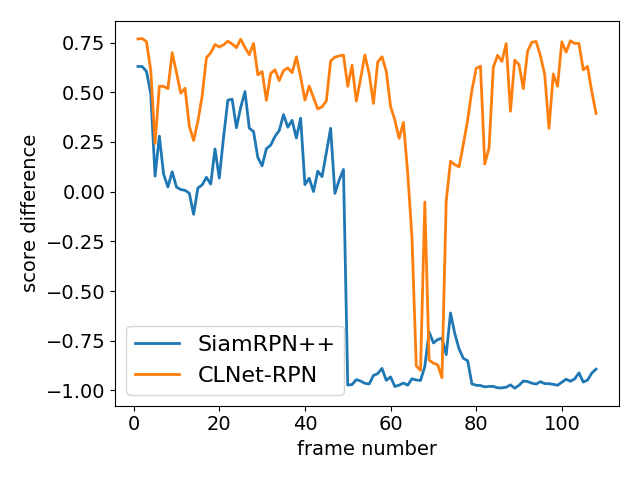}
	\includegraphics[width = .24 \textwidth]{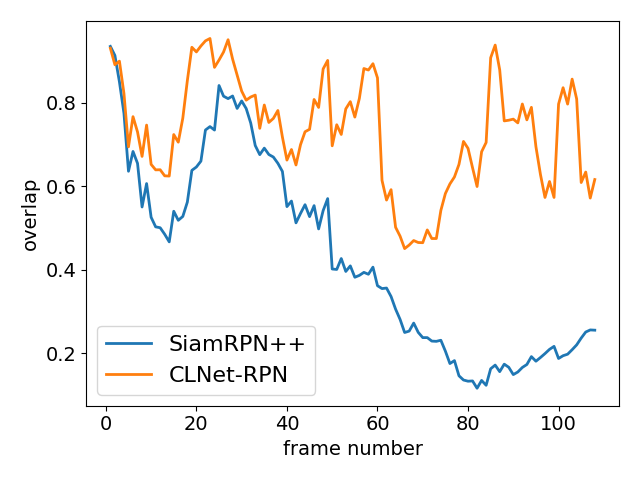}\\
	~~(e) ~~~~~~~~~~~~~~~~~~~~~~~~~~~~~~~~~~~~~~~~~~~(f)
	\caption{\small \textbf{The issue in SiamRPN++~\cite{li2019siamrpn}.} Subfigure (a) shows the decisive bounding boxes (bboxes) provided by SiamRPN++ in two frames of the \textit{SnowBoarding4} sequence in NfS~\cite{kiani_galoogahi2017need}. The blue rectangle represents the decisive positive bbox (P-bbox), which has the highest classification score in the positive set. The blue text shows the classification score ($P_c$) and overlap ($P_o$) with the ground-truth (green rectangle). We also obtain the decisive negative bbox (red rectangle), denoted as N-bbox, and its scores (red text). Similarly, (b) shows the decisive bboxes obtained by our CLNet-RPN. Subfigures (c) and (d) show the decisive classification scores (positive and negative) for the whole sequence, obtained by SiamRPN++ and our CLNet-RPN, respectively. Subfigure (e) plots the score differences for the two methods. Finally, (f) shows the overlap scores of the final tracking bboxes produced by SiamRPN++ and our CLNet-RPN.}
	\label{fig:tracking case}
\end{figure}

\noindent\textbf{Analysis for the Tracking Case.} Firstly, a metric is defined to measure the unsuitable decision hyperplane and better illustrate the issue under the tracking setting. We run the SiamRPN++~\cite{li2019siamrpn} tracker on the \textit{SnowBoarding4} sequence in NfS~\cite{kiani_galoogahi2017need}, and observe the classification scores of alternative bboxes for each frame. Firstly, we group these bboxes into a positive set and a negative set. The bboxes are assigned to the positive set, if their overlap scores with ground-truth bboxes are more than or equal to 0.5. The other bboxes are classified into the negative set. Then, we select the bbox with the highest classification score as the decisive positive bbox (P-bbox) for each positive set. Similarly, we can obtain the decisive negative bbox (N-bbox) from the negative set. As shown in Fig.~\ref{fig:tracking case}(a), we plot the P-bbox (blue rectangle) and N-bbox (red rectangle) of SiamRPN++ for frame 1 and frame 50. We also provide the classification ($P_c \in [0,1]$ and $N_c\in [0,1]$) and overlap ($P_o$ and $N_o$) scores. When the positive set is empty, \textit{i.e.}, the overlaps of all bboxes are less than 0.5, we set $P_c=0$ and $N_c=0$. This also indicates tracking faults.

Then, we can use the score difference value $D = P_c-N_c \in [-1,1]$ to measure the classification decision hyperplane. A large $D$ means the decision hyperplane is discriminative and can separate the samples. Conversely, a small $D$ indicates that the hyperplane is not robust and can easily be misled by some difficult samples. When $D<0$, it faces tracking faults, \textit{i.e.}, the negative bbox ($N_o<0.5$) is regarded as the target bbox. Thus, from this measurement, we find that some unexpected performance can be attributed to an unsuitable decision hyperplane. As shown in Fig.~\ref{fig:tracking case}(c), SiamRPN++ obtains many low score differences before frame 50, resulting in accumulated error. This prevents the model from being able to track the object after frame 50 (see Fig.~\ref{fig:tracking case}(e)).

{Furthermore, we suppose the underlying reason for the unsuitable decision hyperplane is that the tracking model can not see the decisive samples in this sequence (blue and red rectangles in frame \#0001 of Fig.~\ref{fig:tracking case}(a)) during offline training, leading to high positive classification scores for negative samples (Fig.~\ref{fig:tracking case}(c)). To verify this suppose, we adjust the tracking model by using the samples in frame \#0001. As shown in Fig.~\ref{fig:tracking case}(d), we significantly reduce the scores of negative samples and enhance the model's discriminative ability for this sequence. This indicates that the underlying reason for performance degradation can be attributed to the issue of \textit{missing decisive samples} during offline training.
}

\noindent\textbf{Context Ignoring.} We try to alleviate the above issue and seek a reasonable decision hyperplane for tracking. Notice that in the first frame of a sequence, the target ground-truth (GT) bbox is given. It contains rich information for this sequence, which can help generate a reasonable decision hyperplane and enhance the model discrimination ability. However, most Siamese-based trackers ignore the discriminative context information and only apply GT bbox to obtain the template feature. It is vital to effectively utilize the supervision information for the model (classifier) learning. One simple strategy is using the negative and positive samples (i. e. testing data in Fig.~\ref{example}) for finetuning. However, model finetuning is time-consuming because of the requirement of numerous iterations for optimization~\cite{nam2016learning}. Furthermore, the limited number of samples easily results in overfitting. If only the testing data in Fig.~\ref{example} is considered, the optimal decision hyperplane will be $w^2$. However, it will provide the wrong predictions for some difficult training samples. In fact, it is not easy to find the ideal decision hyperplane $w^*$ by considering both the testing and training data.
We try to use a compact latent network (CLNet) to find the ideal model which can distinguish both samples in the first frame (testing data) and large-scale labeled videos (training data).

In our experiments, we find that a discriminative decision hyperplane (with large score difference $D$) can indeed improve the tracking performance. We employ our CLNet (see \S\ref{sec:latent_metanet}) to adjust SiamRPN++ and denote the new tracker as CLNet-RPN. As shown in Fig.~\ref{fig:tracking case}(e)(f), our CLNet achieves larger score difference $D$ than SiamRPN++ in many frames, and also provides better overlaps. Furthermore, we find that $D$ is approximately proportional to the overlap, which indicates that there is an underlying connection between the discriminative decision hyperplane and the tracking performance.
Besides, we also provide two visual examples in Fig.~\ref{fig:tracking case}(b), and plot classification scores ($P_c$ and $N_c$) of decisive bboxes in Fig.~\ref{fig:tracking case}(d). Compared with Fig.~\ref{fig:tracking case}(a)(c), we find that our CLNet significantly improves the discrimination ability for the negative samples, achieving better performance.


\subsection{Compact Latent Network for SiamRPN++}\label{sec:latent_metanet}
The recent state-of-the-art Siamese-based model, SiamRPN++~\cite{li2019siamrpn}, is selected as the base tracker. Then, we apply our CLNet to this basic model to demonstrate its effectiveness, denoting the new tracker as CLNet-RPN. 

All layers are fixed except the last one in the regression and classification branches, to maintain the original model's generality. Note that, we only explain incorporating classification branches with our compact latent network, but the regression branches can be processed with a similar method. We omit the notation $cls$ for simplicity. Firstly, the base model is reformulated to better explain our new module. As mentioned in \S\ref{sec:siamrpn}, SiamRPN++ has two convolutional layers in the last head block, which can be decomposed into two components, i.e. $head=head_1(head_0)$. In the classification branch, we use a function $f$ to denote all layers except the last one and reformulate the classification map in Eq. (\ref{eq:rpn-map}):
\begin{equation}\label{cls-map}
\begin{aligned}
\mathbf{A} & = head_1(\mathbf{M}(\mathbf{x},\mathbf{z});\theta_1),
\end{aligned}
\end{equation}
where $\theta_1$ is the parameter in $head_1$, and $\mathbf{M}(\mathbf{x},\mathbf{z}) = head_0\left(\alpha^{fea}(\phi(\mathbf{x})) \star \alpha^{ker}(\phi(\mathbf{z}))\right)$.
In fact, we view $\mathbf{M}\in \mathbb{R}^{w \times h \times c}$ as a $w \times h$ feature map, and the channel number is $c$. 

A compact feature representation is provided to better utilize the feature map $\mathbf{M}$ via exploiting its statistical information. Firstly, the channel number $c$ is adjusted to $\bar{c}$ to acquire the initial latent feature $\mathbf{\bar{M}}$, by using a network with three $1\times 1$ convolutional layers, called \textbf{feature-adjusting subnetwork} $g_{a}$ (see the details in \S\ref{sec:imp details}). The formulation is as follows:
\begin{equation}\label{eq:adjust subnet}
    \mathbf{\bar{M}}=g_{a}(\mathbf{M}),
\end{equation}
where $\mathbf{\bar{M}}\in \mathbb{R}^{w \times h \times \bar{c}}$. We set $\bar{c} \leq 2c$ to reduce the cost. 

We regard this latent feature $\mathbf{\bar{M}}$ as a set containing the hidden vectors of the template-instance pair, i.e., $\mathbf{\bar{M}}= \{\mathbf{\bar{m}}_1,\mathbf{\bar{m}}_2, \cdots, \mathbf{\bar{m}}_{wh}\}$.
We split this set into two groups according to the classification labels $\mathbf{Y}\in \mathbb{R}^{w \times h}$, i. e. a negative set $\mathcal{N}=\{\mathbf{\bar{m}}^-_1,\mathbf{\bar{m}}^-_2, \cdots, \mathbf{\bar{m}}^-_{n^-}\}$, and a positive set $\mathcal{P}=\{\mathbf{\bar{m}}^+_1,\mathbf{\bar{m}}^+_2, \cdots, \mathbf{\bar{m}}^+_{n^+}\}$.
To capture the statistical information, the \textbf{latent encoder} operation is presented via the standard deviation $\sigma$ and mean $\mu$ of these two sets:
\begin{equation}
\begin{aligned}
\mu^\rho &= \frac{1}{n^\rho}\sum\nolimits_{i=1}^{n^\rho}\mathbf{\bar{m}}^\rho_i, \\
\sigma^\rho &=\sqrt{\frac{1}{n^\rho}\sum\nolimits_{i=1}^{n^\rho}(\mathbf{\bar{m}}^\rho_i-\mu^\rho)^2},
\end{aligned}
\label{eq:mu_std}
\end{equation}
where $\rho\in \{+,-\}$, $\mu\in \mathbb{R}^l$, and $\sigma\in \mathbb{R}+^l$. We use concatenation to build the final latent feature $\mathbf{c}$, i.e.,
\begin{equation}\label{eq:c}
\mathbf{c} = concat(\mu^+,\sigma^+,\mu^-,\sigma^-).
\end{equation}
In summary, we denote our latent encoder operation $LE$:
\begin{equation}\label{eq:LE}
\mathbf{c} = LE(\mathbf{\bar{M}},\mathbf{Y}).
\end{equation}

\begin{figure}[!tb]
	\centering
	\includegraphics[width = .45 \textwidth]{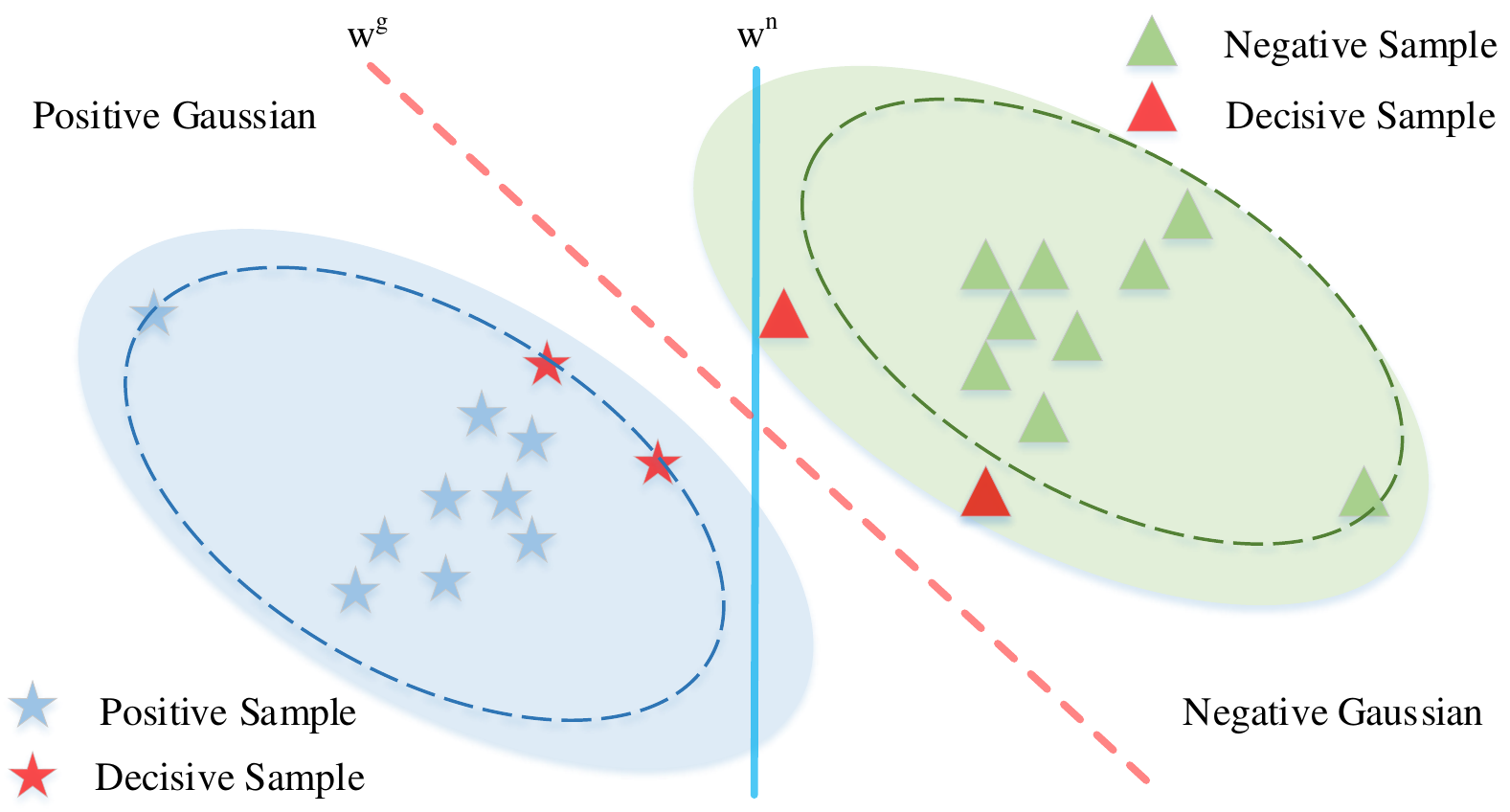}
	\caption{\small {\bf Intuitive example for the benefits of statistics-based features on the binary classification.} $w^n$ and $w^g$ are the decision hyperplanes decided by normal samples and statistics-based features, respectively. Dotted ellipses are the approximations of true Gaussian distributions (blue and green ellipses). }
	\label{example2}
\end{figure}

{
\noindent\textbf{Benefits of Statistics-Based Feature.}
In fact, our statistics-based feature can alleviate the {\it missing decisive samples} issue and is more robust than original features. An intuitive example is given in Fig.~\ref{example2} to verify its powerful representative ability to reduce this issue. We assume the distributions of negative and positive sets are Gaussian distributions (represented by the blue and green ellipses), and draw several negative (green triangles) and positive (blue stars) samples. If only these drawn samples and the rule of largest margin are used to seek the decision hyperplane, a wrong hyperplane $w^n$ will be found. In practice, $w^g$ should be the true decision hyperplane, because the two ellipses' edges are true sample boundaries. The incorrect decision hyperplane can be attributed to {\it missing decisive samples}, \textit{i.e.}, the drawn samples do not include the key samples (red triangle and star). However, if the Gaussian distribution is represented by the statistics features, \textit{i.e.}, the standard deviation and mean of each sample set, it is easier to obtain the approximations (dotted ellipses). Then we can use them to achieve better classification performance and find out the ideal decision hyperplane, by some optimization methods based on uncertainty~\cite{khan2019striking}. In summary, even we do not draw the decisive samples in this intuitive example, the correct decision hyperplane can still be found by using the statistics-based feature.
Besides, note that we only require the mean and standard deviation to represent a Gaussian distribution.
Thus, they can be adopted as the compact features to reduce the {\it missing decisive samples} issue, via unitizing the underlying distribution.
}

Our compact representation has two additional merits. 1) Its parameters are fewer than the original feature. Note that $\bar{c}$ has the same order of magnitude to $c$. In practice, the original feature's parameter number $w\times h\times c$ is usually higher than $4\times \bar{c}$ of the compact feature $\mathbf{c}$. 2) Our compact feature is number-free. This feature can be built by a various number of negative ad positive samples. Since SiamRPN-based tracking models have different input image-pairs~\cite{li2018high}, leading to the changing number of negative or positive samples, our feature is very suitable for these models.

Given the latent compact feature $\mathbf{c}$, we use a multi-layer perceptron to build the \textbf{prediction subnetwork} $g_{\Delta}$ and generate a deviation for the last head layer's weights:
\begin{equation}
\label{eq:pred subnet}
\begin{aligned}
\Delta \theta_1 = g_{\Delta}(\mathbf{c}),
\end{aligned}
\end{equation}
where the weight deviation predictor is $g_{\Delta}$, which including three fully connected layers.
We add these deviations with the weights to tune the model for various sequences:
\begin{equation}
\label{eq:theta_a}
\begin{aligned}
\theta_a = \theta_1+\Delta \theta_1.
\end{aligned}
\end{equation}
Finally, we use the adjusted weight $\theta_a$ to generate the final classification map.
We apply a similar weight adjustment for the regression branch, except that we construct the compact latent feature by using the positive set, because only the positive samples are available during training. Overall, the formulations of adjusted regression map $\mathbf{A}^{loc}_a$ and classification map $\mathbf{A}^{cls}_a$ are as follow:
\begin{equation}
\begin{aligned}
\mathbf{A}^{cls}_a=head_1^{cls}(f^{cls}(\mathbf{x},\mathbf{z});\theta_a^{cls}),\\
\mathbf{A}^{loc}_a=head_1^{loc}(f^{loc}(\mathbf{x},\mathbf{z});\theta_a^{loc}).
\end{aligned}
\label{eq:adjust_maps}
\end{equation}
These maps are used to predict targets, like SiamRPN++~\cite{li2019siamrpn}.

\subsection{Training with Diverse Sample Mining}\label{sec:hard sample}
 All parameters in the original model are fixed to train our compact latent networks, and the same smooth-L1 loss $L^{loc}$ and softmax loss $L^{cls}$ in SiamRPN++~\cite{li2019siamrpn} are used to adjust regression and classification maps, respectively. We formulate the final loss as follows:
\begin{equation}
L = L^{cls}(\mathbf{A}^{cls}_a,y^{cls})+\lambda L^{loc}(\mathbf{A}^{loc}_a,y^{loc}),
\label{eq:loss}
\end{equation}
where $y^{loc}$ and $y^{cls}$ represent the regression and classification ground-truths, respectively, and we set the trade-off parameter $\lambda$ is as $1.2$. To obtain a more discriminative feature, SiamRPN++ samples the image-pairs in one training batch from different sequences. This training method is not suitable for our adjustment. Since our goal is to predict the adaptive weights for one sequence rather than the whole dataset. We do not require the general information across sequences but aim to extract the vital information within a specific sequence during training. Besides, the local statistical information inside one sequence is more suitable for our statistics-based latent feature. Thus, several image-pairs are randomly sampled from the same sequence in each training batch .

A diverse sample mining approach is proposed to enhance the training performance, via ranking unused negative samples with their classification scores. In SiamRPN++~\cite{li2019siamrpn}, adopts the intersection-over-union ($IoU$) between the ground-truth bbox and anchor are adopted to choose the negative and positive samples. When $IoU<0.3$, the sample is negative, and when $IoU>0.6$, it is regarded as positive. One training pair contains 64 samples, where the number of positive samples is at most 16. The diverse samples may exist in the unused negative samples, and they will benefit the discrimination ability of the model. The original score map $\mathbf{A}^{cls}$ is used to mine these diverse samples. According to the positive classification scores of $\mathbf{A}^{cls}$, the unused negative samples are sorted in descending order. We regard the first 16 negative samples as diverse samples and add them to each training pair. Therefore, the total sample number in one training pair is 80. The additional diverse samples are usually near the original model's decision hyperplane and have a high possibility to be the decisive samples. Notice that these samples are helpful for building robust compact latent features and finding the ideal decision hyperplane (See the analysis in \S\ref{sec:latent_metanet}).

\subsection{Online Tracking with Conditional Updating}\label{sec:cu}
{In our previous work \cite{dong2020clnet}, we only run the proposed CLNet in the first frame to capture the scene information and generate adjusting parameters for the base tracker. Then we run the adjusted tracker without any more adjustments to reduce computation load. However, this strategy cannot handle the videos with huge scene variation. Thus, we propose a simple and efficient updated method, called conditional updating (CU). }

{First, we need to select reliable samples as the candidates for updating, since we do not have access to the ground-truth of other frames. Specifically, a tracker usually predicts a bbox with the confidence (classification) score for each frame. If the bbox in a frame has a high score, this indicates that the target in the current frame has a similar appearance with the template (initial target) and the predicted bbox is more likely correct. If not, this means that the current target may suffer from huge appearance variation or occlusion. Extracting samples from this frame can introduce error information for the positive samples. Thus, we only use the bbox with high confidence to generate reliable samples. For simplification, we only provide the formulations for the classification branch. We give a constant threshold $\tau^r$ to select reliable frames and a candidate set $\{\mathbf{M}_c, \mathbf{b}_c\}$ to save the feature map and bbox. We denote the feature map in Eq.~\ref{cls-map} for $i$-th frame as $\mathbf{M}_i$, then the candidate set is defined as
\begin{equation}
    \label{eq:candidate}
    \{\mathbf{M}_c, \mathbf{b}_c\}=
    \left \{\begin{array}{ll}
    \{\emptyset, \emptyset\},     & i=1,  \\
    \{\mathbf{M}_c, \mathbf{b}_c\},     &i>1, s_i\leq\tau_r, \\
    \{\mathbf{M}_i, \mathbf{b}_i\},  &i>1, s_i>\tau_r,
    \end{array}\right.
\end{equation}
where $s_i$ is the confidence score of predicted bbox $b_i$.
}

{
Second, we decide when to update the tracking model. To capture the hard cases caused by huge scene variation or similar distractors, we propose a simple metric based on the classification scores. As mentioned in the visual analysis of \S\ref{sec:analysis}, when the highest negative classification score is larger than the positive one, the tracking fault will occur.
We utilize the margin between the highest positive and negative scores to measure the tracking difficulty for each frame.
Since we do not have the access to the ground-truth bbox on other frames except the first frame, we can only use the predict bbox to generate pseudo classification labels $\mathbf{\bar{Y}_i}$ (We adopt the same label generation methods from the base trackers). According to these labels, we split the classification scores into positive and negative sets.
Then we can denote the margin of $i$-th frame as $\eta_i = s_{p,i}^*-s_{n,i}^*$,
where $s_{p,i}^*$, $s_{n,i}^*$ are highest scores from positive and negative sets, respectively. Small margin means the samples in the current frame and adjacent frame are difficult to distinguish by using the current tracking model. We need to adjust the tracking model for better performance in the next frame. Thus, we set a constant threshold $\tau_m$ to decide whether updating the tracking model, \textit{i.e.}, when $\eta_i<\tau_m$, we update the tracking model.
}
{To avoid updating frequently, we clean the candidate set, \textit{i.e.}, $\{\mathbf{M}_c, \mathbf{b}_c\}=\{\emptyset, \emptyset\}$ after updating once. We stop to update model until we find a new reliable frame and the score margin is less than the threshold $\tau_m$.
}

To distinguish the original CLNet in \cite{dong2020clnet}, we denote our new proposed model with conditional updating as CLNet*.

\subsection{Extensions to Other Siamese-Based Trackers}\label{sec:extension}
We apply our CLNet* to two other representative Siamese-based trackers: SiamFC~\cite{bertinetto2016fully-convolutional} and SiamBAN~\cite{chen2020siamese}.
To distinguish the different trackers adjusted by our method, we denote them as CLNet*-RPN, CLNet*-FC, and CLNet*-BAN, which correspond to SiamRPN++, SiamFC, and SiamBAN, respectively.

\noindent{\bf CLNet*-FC.} As mentioned in \S\ref{sec:analysis}, for SiamFC, the template feature $\phi(\mathbf{z})$, and offset $b$ in Eq. (\ref{siamfc}), in SiamFC can be regarded as the classifier for discriminating the instance patch in the search region. Thus, we can use our CLNet to adjust the template feature and the offset. In SiamFC, the final similarity map $\mathbf{S}$ in Eq.~\ref{siamfc} is regarded as the feature map (like $\mathbf{M}$ in Eq.~\ref{cls-map}). This is the input of our CLNet, since we require the correlation features between the instances and template to build our CLNet and only the similarity map contains the correlation features in SiamFC. We denote the feature map as $\mathbf{M}_{f} = \mathbf{S}$. According to the classification labels, we can extract the latent compact feature $\mathbf{c}_{f}$ via Eq.~\ref{eq:adjust subnet}, \ref{eq:mu_std} and \ref{eq:c} in \S\ref{sec:latent_metanet}. A prediction subnetwork $g_{\Delta_f}$ is used to produce the deviations for the template feature and offset:
\begin{equation}
    [\Delta\phi(\mathbf{z}),\Delta b]=g_{\Delta_f}(\mathbf{c}_{f}),
\end{equation}
where $g_{\Delta_f}$ is a deviation predictor that includes three fully connected layers, similar to $g_{\Delta}$ in Eq.~\ref{eq:pred subnet}. We add these deviations into the corresponding template feature and offset:
\begin{equation}
    \phi(\mathbf{z})_a=\phi(\mathbf{z})+\Delta\phi(\mathbf{z}),~~~b_a=b+\Delta b.
\end{equation}
Finally, the adjusted similarity map $\mathbf{S}_f$ is formulated as:
\begin{equation}
    \mathbf{S}_f = \phi(\mathbf{x})* \phi(\mathbf{z})_a + b_a.
\end{equation}
Similar to SiamFC, the adjusted map is applied to predict the target. Notice that SiamFC uses all samples for training. Thus, we omit the diverse sample mining (\S\ref{sec:hard sample}) in our CLNet*-FC.

\noindent{\bf CLNet*-BAN.} The recent SiamBAN removes the pre-defined anchors to obtain faster and better performance than its baseline SiamRPN++. Thus, the architecture of SiamBAN is very similar to SiamRPN++, and only the head part has slight differences. To obtain CLNet*-BAN, we only need to change the number of output channels in the prediction subnetwork of our CLNet. The other adjustment procedures and training method are the same as CLNet*-RPN, described in \S\ref{sec:latent_metanet} and \ref{sec:hard sample}. The detailed structures of CLNets can be found in \S\ref{sec:imp details}.

\begin{table}
\centering
\caption{\small
\textbf{Size of output feature map in each layer of our CLNets.} This table provides the detailed structures of the CLNets inside our CLNet*-FC and CLNet*-RPN trackers, which are based on SiamFC-Pt~\cite{bertinetto2016fully-convolutional}, and SiamRPN++~\cite{li2019siamrpn}, respectively.
H, W, and C represent the height, weight, and channel number of the feature map, respectively.
$\bar{c}$, $\dot{c}$, and $k$ are set to 128, 256 and 5, respectively.}
	\label{tab:structures}
\setlength{\tabcolsep}{0.8mm}{
\begin{tabular}{ccc|cc|cc}
\hline
          & \multicolumn{2}{c|}{CLNet*-FC}       & \multicolumn{4}{c}{CLNet*-RPN}  \\
          \hline
          & \multicolumn{2}{c|}{Classification} & \multicolumn{2}{c|}{Classification} & \multicolumn{2}{c}{Regression} \\
          \hline
          & HxW             & C                & HxW          & C                   & HxW        & C                 \\
          \hline
Input     & 25x25           & 1                & 25x25        & 256                 & 25x25      & 256               \\
\hline
Conv1     & 25x25           & $\bar{c}$              & 25x25        & $\bar{c}$                 & 25x25      & 2*$\bar{c}$               \\
Conv2     & 25x25           & $\bar{c}$              & 25x25        & $\bar{c}$                 & 25x25      & 2*$\bar{c}$               \\
Conv3     & 25x25           & $\bar{c}$              & 25x25        & $\bar{c}$                 & 25x25      & 2*$\bar{c}$               \\
\hline
$LE$ & 1x1             & 4*$\bar{c}$            & 1x1          & 4*$\bar{c}$               & 1x1        & 4*$\bar{c}$             \\
\hline
FC1       & 1x1             & $\dot{c}$              & 1x1          & $\dot{c}$                 & 1x1        & $\dot{c}$               \\
FC2       & 1x1             & $\dot{c}$              & 1x1          & $\dot{c}$                 & 1x1        & $\dot{c}$               \\
FC3       & 1x1             & 25*25+1          & 1x1          & 2k*(256+1)+1        & 1x1        & 4k*(256+1)+1 \\
\hline
\end{tabular}
}

\end{table}

\section{Experimental Results}\label{sec:experiment}
{Our algorithm is implemented in Pytorch (Python 3) and evaluated on a single GPU (NVIDIA RTX 2080Ti).}
We use different versions of Pytorch according to the base trackers. For CLNet*-FC, we select the Pytorch-version\footnote{\url{https://github.com/StrangerZhang/SiamFC-PyTorch}} of SiamFC~\cite{bertinetto2016fully-convolutional} with Pytorch 0.4.0 as the base code. This base tracker is denoted as SiamFC-Pt. For CLNet*-RPN and CLNet*-BAN, we use the official codes from SiamRPN++~\cite{li2019siamrpn} and SiamBAN~\cite{chen2020siamese}, and develop our method in the recommended Pytorch 0.4.0 and Pytorch 1.3.1, respectively.
We compare our trackers, CLNet*-RPN, CLNet*-FC, and CLNet*-BAN with several representative trackers on NfS~\cite{kiani_galoogahi2017need}, DTB~\cite{li2017visual}, LaSOT~\cite{fan2019lasot}, GOT10k~\cite{huang2019got-10k}, VOT2019~\cite{Kristan2019a}, and VOT2020~\cite{Kristan2020a}.
{The tracking speeds of CLNet*-RPN,  CLNet*-FC, and CLNet*-BAN are 38.1 FPS, 104.9 FPS, and 43.6 FPS.}

\subsection{Implementation Details}\label{sec:imp details}
\subsubsection{Architecture}
{The proposed CLNet includes a feature-adjusting subnetwork $g_{a}$, one latent encoder operation $LE$, and a prediction subnetwork $g_{\Delta}$. $g_{a}$ includes three convolutional blocks (named as Conv1, Conv2, and Conv3), which are built by a $1 \!\times\! 1$ convolution, batch-norm, and ReLU layer. $g_{\Delta}$ includes three fully connected layers (denoted FC1, FC2, and FC3), where the first two layers are followed by a ReLU layer and the last one is followed by a \textit{Tanh} layer.}
The settings of \textbf{CLNet*-RPN} and \textbf{CLNet*-FC} are shown in Table~\ref{tab:structures}. For \textbf{CLNet*-BAN}, we only need to set the anchor number $k=1$ in Table~\ref{tab:structures}.

\subsubsection{Training}
We follow the training protocol of the basic trackers: SiamRPN++, SiamFC-Pt, and SiamBAN, using the same training datasets.
{Since we want to eliminate the impact of different training data for each basic tracker, and explore whether our approach can further improve each basic tracker with the same training data.}
{In fact, we freeze the networks in the basic trackers and only train the CLNets inside them. Thus, the number of training iterations is less than that of the basic trackers (1/10 number of SiamRPN++ and SiamBAN)}. 

\noindent{\textbf{CLNet*-RPN}.}
{COCO~\cite{lin2014microsoft}, DET~\cite{russakovsky2015imagenet}, VID~\cite{russakovsky2015imagenet}, and YouTubeBB~\cite{real2017youtube-boundingboxes} are adopted as training datasets. We use Synchronized stochastic gradient descent (SGD) over four GPUs. The batch size is 64. Each epoch contains 60,000 training pairs. We train over 20 epochs, where the first five epochs are used for warrmup with a step learning rate from 0.001 to 0.005, and we exponentially decay the learning rate from 0.005 to 0.0005 in the last 15 epochs. 
The other hyperparameters are the same as for SiamRPN++.}

\noindent{\textbf{CLNet*-FC}.}
We use the VID~\cite{russakovsky2015imagenet} dataset and train our network over 30 epochs with SGD on a single GPU. In fact, the training protocol and the hyperparameters are the same as for the basic tracker, \textit{i.e.}, the Pytorch-version of SiamFC.

\noindent{\textbf{CLNet*-BAN}.}
The training datasets includes COCO~\cite{lin2014microsoft}, DET~\cite{russakovsky2015imagenet}, VID~\cite{russakovsky2015imagenet}, YouTubeBB~\cite{real2017youtube-boundingboxes},  GOT10k~\cite{huang2019got-10k} and LaSOT~\cite{fan2019lasot}. Our network is trained with SGD over four GPUs. The batch size is 112 (28 per GPU). We train over 20 epochs, each of which contains 10,000 training pairs. The other hyperparameters are the same as for SiamBAN.

\subsection{Comparison on the NfS30 Dataset}\label{sec:nfs}
\begin{figure}[!tb]
	\centering	
	\includegraphics[width = .24\textwidth]{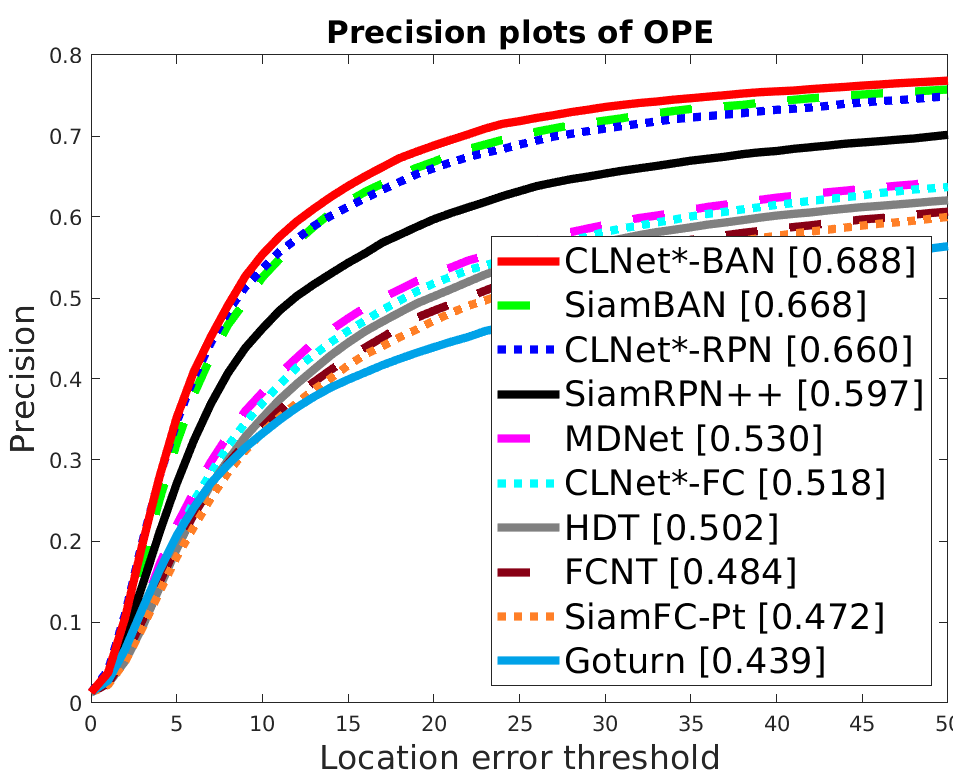}
	\includegraphics[width = .24\textwidth]{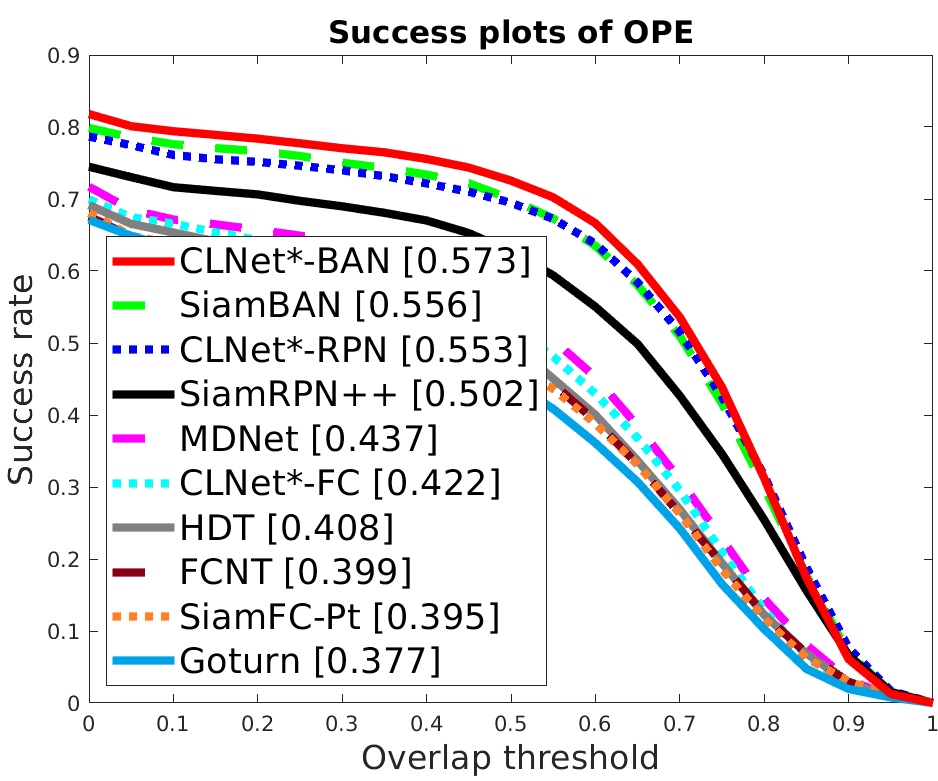}	
	\caption{\small \textbf{Precision and overlap success plots with AUC on the NfS30 dataset~\cite{kiani_galoogahi2017need}.} Only the top-ten trackers are shown.}
	\label{fig:nfs}
\end{figure}

Need for Speed (NfS)~\cite{kiani_galoogahi2017need} includes 100 challenging sequences with fast-moving objects. We evaluate the basic models, SiamRPN++~\cite{li2019siamrpn}, SiamFC-Pt~\cite{bertinetto2016fully-convolutional}, and SiamBAN~\cite{chen2020siamese}, and our trackers, CLNet*-RPN, CLNet*-FC, and CLNet*-BAN, on the 30 FPS version (NfS30). We also compare with the several well-known trackers evaluated in \cite{kiani_galoogahi2017need}, including Correlation Filter (CF) based trackers (SRDCF~\cite{danelljan2015learning}, DSST~\cite{danelljan2014adaptive}, KCF~\cite{henriques2015high-speed}, LCT~\cite{ma2015long-term}, SAMF~\cite{li2014scale}, HCF~\cite{ma2015hierarchical}, and HDT~\cite{qi2016hedged}), and deep trackers (MDNet~\cite{nam2016learning}, FCNT~\cite{wang2015visual}, and GOTURN~\cite{held2016learning}).
As shown in Fig.~\ref{fig:nfs}, our CLNet*-BAN achieves the best performance in terms of precision and AUC, which are $29.8\%$ and $31.1\%$ higher than the best tracker, MDNet~\cite{nam2016learning}, reported in the original paper. Compared with the SiamBAN, our tracker also obtains promising performance improvement in both metrics.
Furthermore, our CLNet*-RPN achieves significant relative gains of $10.6\%$ and $10.2\%$ in terms of both precision and area-under-the-curve (AUC), compared with the baseline SiamRPN++. Besides, the proposed CLNet*-FC ranks second among the original trackers in \cite{kiani_galoogahi2017need}. Notably, it significantly outperforms its baseline SiamFC-Pt, by large relative gains of $9.7\%$ and $6 .8\%$ in terms of precision and AUC.

\subsection{Results on the DTB70 Dataset}\label{sec:dtb}
\begin{figure}
	\centering	
	\includegraphics[width = .24\textwidth]{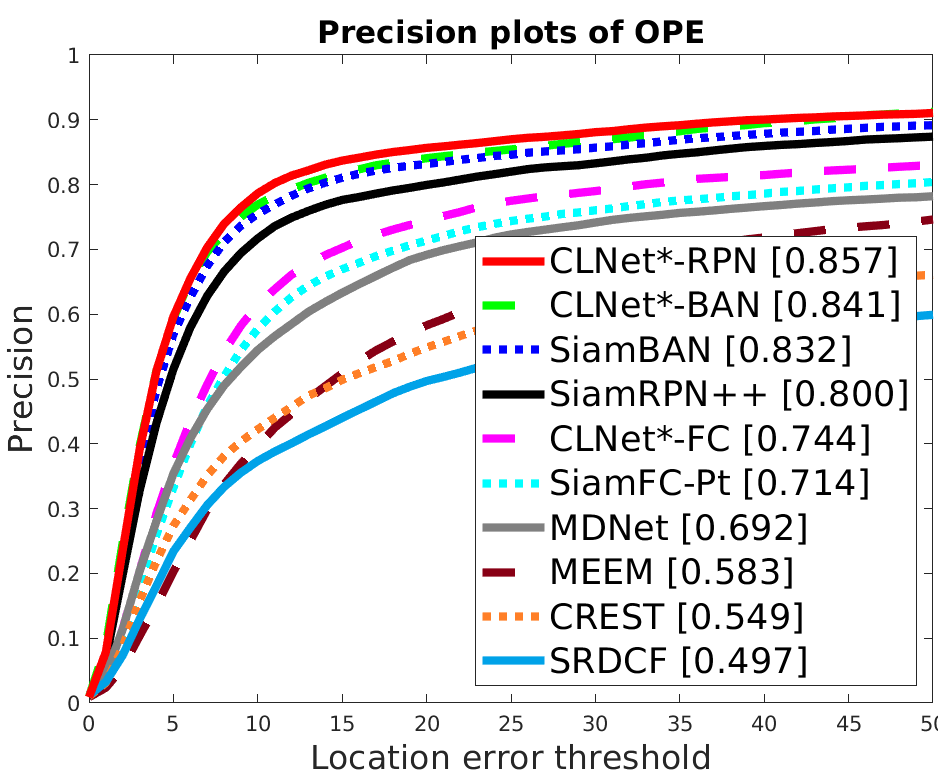}
	\includegraphics[width = .24\textwidth]{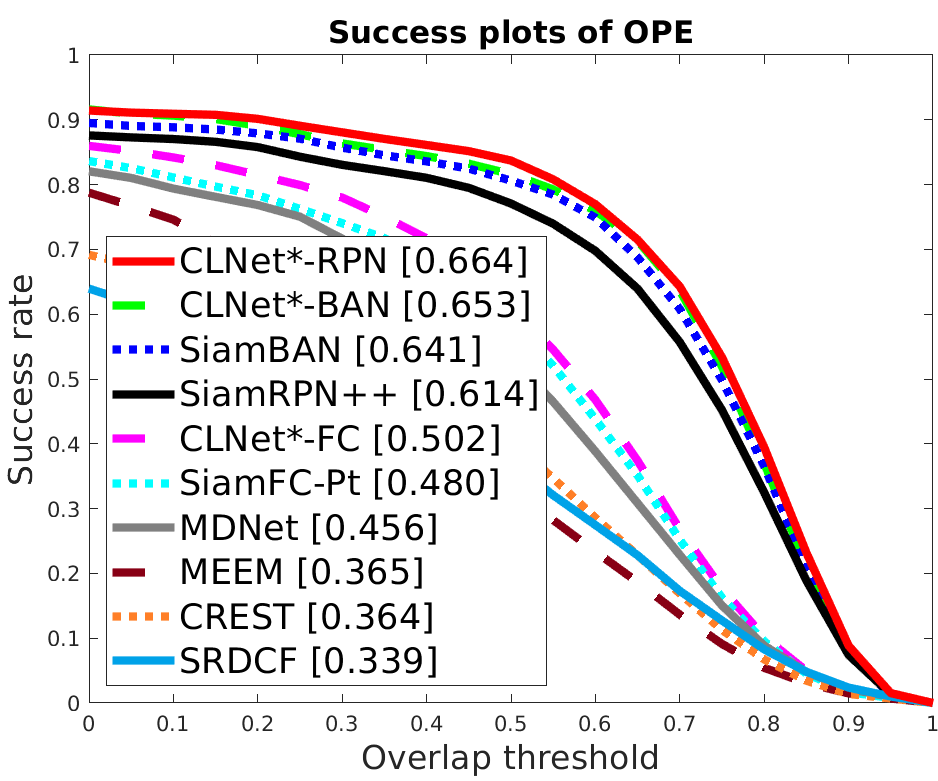}	
	\caption{\small \textbf{Precision and overlap success plots with AUC on the DTB70 dataset~\cite{li2017visual}.} Only the top-ten trackers are shown.}
	\label{fig:dtb}
\end{figure}

Our trackers are also evaluated on the DTB Dataset~\cite{li2017visual}, which includes 70 videos.
We report the precision and AUC plots in Fig.~\ref{fig:dtb}, comparing against the basic trackers, as well as other trackers reported in DTB70, such as MDNet~\cite{nam2016learning}, CREST~\cite{song2017crest}, MEEM~\cite{Zhang2014MEEM}, and SRDCF~\cite{danelljan2015learning}. 
{Among all the trackers, our CLNet*-RPN is the best tracker, which outperforms the basic tracker, SiamRPN++, by large gains of $7.1\%$ and $8.1\%$ in terms of precision and AUC.
In both evaluation metrics, the proposed CLNet*-BAN tracker obtains the second-best score, achieving superior performance to its baseline SiamBAN. Moreover, our CLNet*-FC also improves its baseline SiamFC-Pt with promising gains in precision and AUC, and achieves $7.5\%$ and $10.1\%$ higher performance than the best MDNet tracker, reported in the original benchmark~\cite{li2017visual}.
We attribute the performance promotion to the proposed compact latent network.}

\begin{table*}
\centering
\caption{\small
\textbf{Comparison on the LaSOT dataset~\cite{fan2019lasot}.} The \red{first} and \blue{second} best scores are highlighted in red and blue, respectively.
}
	\label{res:lasot}
\setlength{\tabcolsep}{0.5mm}{
\begin{tabular}{ccccccccccccccc}
    \toprule
        ~ & MLT & GradNet & SiamDW & C-RPN & PrDiMP & SiamRCNN & Ocean & SiamFC-Pt & CLNet*-FC & SiamRPN++ & CLNet*-RPN & SiamBAN & CLNet*-BAN \\
        ~ & \cite{choi2019deep} & \cite{li2019gradnet} & \cite{zhang2019deeper} & \cite{fan2019siamese} & \cite{danelljan2020probabilistic} & \cite{voigtlaender2020siam} & \cite{zhang2020ocean} & \cite{bertinetto2016fully-convolutional} & \bf Ours & \cite{li2019siamrpn} & \bf Ours & \cite{chen2020siamese} & \bf Ours\\ \midrule
        P ($\%$)~$\uparrow$ & - & 35.1 & - & 44.3 & - & - & \red{52.6} & 32.1 & 33.3 & 48.5 & 50.1 & 51.9 &\blue{ 52.9} \\
        P$_{norm}$ ($\%$)~$\uparrow$ & - & - & - & 54.2 & - & \red{72.2} & - & 41.2 & 42.4 & 56.5 & 58.3 & {59.7} & 62.3 \\
        AUC ($\%$)~$\uparrow$ & 34.5 & 36.5 & 38.4 & 45.5 & \blue{59.8} & \red{64.8} & 52.6 & 33.2 & 33.3 & 49.3 & 50.2 & {51.4} & 52.6 \\ \bottomrule
    \end{tabular}
}
\end{table*}

\begin{table*}[]
\centering
\caption{\small
\textbf{Comparison on the GOT10k dataset~\cite{huang2019got-10k}}. The \red{first} and \blue{second} best scores are highlighted in red and blue, respectively.
}
	\label{res:got10k}
\setlength{\tabcolsep}{1mm}{
\begin{tabular}{cccccccccccc}
\toprule
           & MDNet & GOTURN & MemTracker & PrDiMP & SiamRCNN & SiamFC-Pt & CLNet*-FC & SiamRPN++ & CLNet*-RPN & SiamBAN & CLNet*-BAN \\
           &\cite{nam2016learning}  &  \cite{held2016learning}      &  \cite{yang2018learning}    & \cite{danelljan2020probabilistic}   &\cite{voigtlaender2020siam}        & \cite{bertinetto2016fully-convolutional} & \bf Ours      &   \cite{li2019siamrpn}          &\bf Ours      & \cite{chen2020siamese}         &\bf Ours       \\
\midrule
AO (\%)    & 35.2  & 41.8   & 46.0       & \blue{63.4}     & \red{64.9}     & 34.0      & 35.0     & 51.7      & 54.3      & 55.9    & 56.5      \\
SR0.5 (\%) & 36.7  & 47.5   & 52.4       & \red{73.8}     & \blue{72.8}     & 37.4      & 38.5     & 61.6      & 64.4      & 66.5    & 67.5     \\
\bottomrule
\end{tabular}
}
\end{table*}

\begin{table*}
\centering
\caption{\small
\textbf{Results of the real-time setting in VOT2019~\cite{Kristan2019a}. }
		The \red{first} and \blue{second} best scores are highlighted in red and blue, respectively.
  }
	\label{res:VOT2019}
\setlength{\tabcolsep}{1mm}{
\begin{tabular}{ccccccccccccc}
\toprule
    & SPM   & SiamMask & SiamDW\_ST & DiMP  & SiamFCOT & SiamMargin & SiamFC-Pt & CLNet-FC & SiamRPN++ & CLNet-RPN & SiamBAN & CLNet-BAN \\
    & \cite{wang2019spm-tracker}     & \cite{wang2019fast}         &\cite{zhang2019deeper}            & \cite{bhat2019learning}      &  \cite{Kristan2019a}        &   \cite{Kristan2019a}          & \cite{bertinetto2016fully-convolutional}       &\bf Ours     &    \cite{li2019siamrpn}       &\bf Ours      & \cite{chen2020siamese}        &\bf Ours      \\
    \midrule
EAO~$\uparrow$  & 0.275 & 0.287    & 0.299      & 0.321 & 0.350    &\red{0.366}      & 0.180   & 0.192    & 0.285     & 0.313     & 0.322   &\blue{0.356}     \\
ACC~$\uparrow$  & 0.577 & 0.594    & 0.600      & 0.582 & \blue{0.601}    & 0.585      & 0.511  & 0.507    & 0.599     & \red{0.606}     & 0.596   & 0.588     \\
ROB~$\downarrow$  & 0.507 & 0.461    & 0.467      & 0.371 & 0.386    & \red{0.321}      & 0.968  & 0.883    & 0.482     & 0.461     & 0.401   & \blue{0.341}   \\
\bottomrule
\end{tabular}
}
\end{table*}

\subsection{Results on Large-Scale Datasets}\label{sec:lasot}
\noindent\textbf{LaSOT.}
LaSOT~\cite{fan2019lasot} includes 1,400 sequences with an average sequence length of 2,512 frames. We conduct experiments on the testing set of LaSOT with 280 sequences, {comparing with the recent MLT~\cite{choi2019deep}, GradNet~\cite{li2019gradnet}, SiamDW~\cite{zhang2019deeper}, C-RPN~\cite{fan2019siamese}, PrDiMP~\cite{danelljan2020probabilistic}, SiamRCNN~\cite{voigtlaender2020siam}, and offline Ocean~\cite{zhang2020ocean}.}
Following \cite{fan2019lasot}, three metrics are used for evaluation, including the overlap success (AUC), precision (P), and normalized precision (P$_{norm}$).
As shown in Table~\ref{res:lasot}, all of our trackers achieve better performance than their baselines in terms of all three metrics. Especially, CLNet*-BAN outperforms the baseline SiamBAN with a large relative gain of $4.9\%$ in terms of P$_{norm}$. SiamRCNN and PrDiMP outperform our best tracker CLNet*-BAN, since they have well-designed online learning approaches. While our CLNet*-BAN obtains faster speed around 43.6 FPS, compared with 4.6 FPS of SiamRCNN and 30 FPS of PrDiMP (provided by original papers).

\noindent\textbf{GOT10k.}
{GOT10k~\cite{huang2019got-10k} contains more than 10,000 videos, convering 640+ generic classes. We conduct experiments on the testing set with 180 videos (127 frames average length), comparing with MDNet~\cite{nam2016learning}, GOTURN~\cite{held2016learning}, MemTracker~\cite{yang2018learning}, PrDiMP~\cite{danelljan2020probabilistic}, and SiamRCNN~\cite{voigtlaender2020siam}. We use the average overlap (AO) and success rate (SR0.5) for evaluation. As shown in Table~\ref{res:got10k}, our trackers outperform baselines in both two metrics. Notice that, CLNet*-BAN achieves significant relative gains of 5.0\% and 4.5\% in terms of AO and SR0.5. Our best tracker CLNet*-BAN ranks third in both AO and SR0.5, which is lower than PrDiMP and SiamRCNN. While our running speed is faster than them as mentioned before.}

\subsection{Comparison on the VOT Datasets}\label{sec:vot}
To measure the short-term tracking performance, we also evaluate our trackers on the real-time challenge provided by VOT2019~\cite{Kristan2019a} and VOT2020~\cite{Kristan2020a}. VOT2019 performs re-initialization of a tracker when it fails to track the target, which is the same as the previous VOT challenge. VOT2020 proposes a new anchor-based short-term tracking evaluation protocol by replacing the re-initialization with anchor-based initialization, \textit{i.e.}, placing several anchors on each frame as initialization points. Notice that under the VOT protocol, we will re-initialize our tracker by using CLNet many times. To reduce time cost, we omit the conditional updating strategy (CU), since we have updated our model at each initialization point. We denote our trackers without CU as CLNet-FC, CLNet-RPN, and CLNet-BAN for three basic trackers.

\noindent\textbf{VOT2019.}
{We adopt the EAO, ACC and ROB {under the real-time setting} to compare different trackers, including our basic trackers, the top-two algorithms (SiamMargin and SiamFCOT) ranked by EAO in the real-time challenge, and several recent methods: SPM~\cite{wang2019spm-tracker}, SiamMask~\cite{wang2019fast}, SiamDW\_ST~\cite{zhang2019deeper}, and DiMP~\cite{bhat2019learning}.
As shown in Table~\ref{res:VOT2019}, our CLNet-RPN obtains the best score in terms of ACC. Compared with its baseline SiamRPN++, the proposed method achieves a significant performance improvement of $9.8\%$ in terms of EAO, and also boosts the ACC and ROB scores.
Our CLNet-BAN performs the second best in terms of EAO and ROB among all trackers, and improves its baseline, SiamBAN, by a large gain of $10.6\%$ in terms of EAO. The last tracker, CLNet-FC, also boosts the EAO and ROB scores of its baseline, SiamFC-Pt. These results demonstrate that our approaches can effectively improve the short-term tracking ability of the base models.}

\noindent\textbf{VOT2020.}
VOT2020 re-defines the accuracy and robustness according to the anchor-based initialization. Here, we denote them as A and R. Firstly, basic trackers are evaluated under the real-time setting. As shown in Table~\ref{res:VOT2020}, all of them obtain lower EAO scores than VOT2019, which indicates that new evaluation is more challenging. Then we run our trackers and achieve better EAO scores than all basic trackers. This demonstrates the effectiveness under the new evaluation.
Our models obtain slight improvement compared with the VOT2019 results, since VOT2020 adopts segmentation masks for evaluation, while the baselines only produce bounding boxes leading to our CLNet cannot provide proper adjustments.
Besides there is still a gap between our trackers and the champion AlphaRef~\cite{Kristan2020a} (0.486 EAO) since our basic trackers do not combine any mask refine strategy.
We also provide the details of parameters and Multiply-accumulate Operations for all models. This indicates that our models only take a few additional space and computation costs while achieving promising improvement.

\begin{table}[]
\caption{\small
\textbf{Results of the real-time setting in VOT2020~\cite{Kristan2020a}. }
		The \red{first} and \blue{second} best scores are highlighted in red and blue, respectively. Pm and MAC are Parameters and Multiply-accumulate Operations, respectively.
  }
	\label{res:VOT2020}
\centering	
\resizebox{0.5\textwidth}{!}{
\setlength{\tabcolsep}{0.3mm}{
\begin{tabular}{ccccccc}
\toprule
    & SiamFC-Pt & CLNet-FC & SiamRPN++ & CLNet-RPN & SiamBAN & CLNet-BAN \\
    &\cite{bertinetto2016fully-convolutional}       &\bf Ours     &    \cite{li2019siamrpn}       &\bf Ours      & \cite{chen2020siamese}        &\bf Ours         \\
    \midrule
EAO~$\uparrow$ & 0.172     &  0.177        & 0.242     & 0.261     & \blue{0.263}   & \red{0.266}     \\
A~$\uparrow$ & 0.422      &  0.423        & 0.442     &\red{ 0.458}     & \blue{0.449}   & 0.446     \\
R~$\uparrow$ & 0.479      & 0.499         & 0.678     & 0.69      & \blue{0.714}   & \red{0.722}    \\
\hline
Pm (M) & 2.336 & 3.632 & 53.951 & 61.823 & 53.932 & 57.086 \\
MACs (G)   & 3.190 & 3.201 & 59.521 & 60.009 & 59.509 & 60.004 \\
\bottomrule
\end{tabular}
}
}
\end{table}

\subsection{Ablation Study}\label{sec:ablation}
{The analysis experiments are conducted on a combined dataset including NfS30 (30 FPS version)~\cite{kiani_galoogahi2017need} and DTB70~\cite{li2017visual} datasets. The new dataset includes 170 videos for thorough analysis. The baseline and variants of our approach are evaluated using AUC and precision (P) metrics~\cite{wu2013online}. }

\noindent\textbf{Analysis of Key Components.}
{To explore the influence of additional training iterations,
we firstly retrain the base tracker ({\bf BT}), SiamRPN++, by using our training method, and evaluate it on the combined dataset. As shown in Table~\ref{res:ablation}, the retrained tracker ({\bf RT}) slightly improve AUC ({\bf BT}: 55.8\% vs {\bf RT}: 56.1\%) and precision ({\bf BT}: 69.9\% vs {\bf RT}: 71.4\%). This indicates it will not provide significant gains by simplely using additional training iterations.
Furthermore, to analyze the influence of our key components, they are added to our baseline ({\bf BT}) one by one.
We first use the adjustment network removing the latent encoder to tune the base model ({\bf +AN}).
The gains in terms of AUC ($1.2\%$) and P ($1.1\%$) scores demonstrate that the adjustment network can extract sequence-specific information to improve the base model.
When adding the latent encoder ({\bf +LE}), our CLNet obtains further improvement with a AUC gain of $1.9\%$ and P gain of $1.5\%$. This indicates the advantages of our latent encoder in extract a compact and effective feature containing the sequence-specific information.
Furthermore, the image-pairs are sampled from one sequence in each batch ({\bf +OS}) to obtain a robust latent feature. This achieves a further improvement, with $0.2\%$ and $1.3\%$ gains in terms of AUC and P scores, respectively.
Our diverse sample mining technique ({\bf +DM}) increases AUC and P scores by another $0.9\%$ and $0.6\%$. Finally, the conditional updating ({\bf +CU}) strategy is applied for online tracking, and provide a AUC gain of $1.0\%$ and P gain of $1.3\%$. In summary, compared with our baseline, the final version ({\bf +CU}) achieves significant relative gains of $9.3\%$ and $8.3\%$ in terms of AUC and P scores.}
Besides, to investigate the impact on the regression branch, we can remove CLNets of the regression branch. We find that only adjusting the classification branch ({\bf -RB}) obtain lower scores with $1.0\%$ and $1.3\%$ reduction in terms of AUC and P scores, compared with adjusting both branches ({\bf +DM}). The results indicate that our CLNet can also provide benefits for the regression task, demonstrating the generalization.

\begin{table}
	\centering
		\caption{\small
		{\textbf{Analysis of key components with AUC and precision (P) on combined NfS30~\cite{kiani_galoogahi2017need} and DTB70~\cite{li2017visual} dataset.} The base tracker ({\bf BT}) is SiamRPN++~\cite{li2019siamrpn}. {\bf RT} represents the retrained tracker. The components contain an adjustment network ({\bf +AN}), latent encoder ({\bf +LE}), one sequence for training ({\bf +OS}), diverse sample mining ({\bf +DM}), and conditional updating ({\bf +CU}). {\bf -RB} means omitting the regression branch.}
		}
	\label{res:ablation}
	\setlength{\tabcolsep}{2mm}{
		\begin{tabular}{ccccccccc}
			\toprule
			 & RT & BT & +AN & +LE & +OS & +DM & -RB & +CU\\
			\midrule

		P ($\%$)   & 71.4 & 69.9 & 71.0 & 72.5 & 73.8 & 74.4 & 73.1 & 75.7 \\
		AUC ($\%$) & 56.1 & 55.8 & 57.0 & 58.9 & 59.1 & 60.0 & 59.0 & 61.0  \\
			\bottomrule
		\end{tabular}
	}

\end{table}

\noindent\textbf{Different Weight Augmentation.}
{
As mentioned before, our CLNet produces a deviation of the weights of the last layers (Eq.~\ref{eq:pred subnet}) and we use a simple additive augmentation to fuse it and the original weights (Eq.~\ref{eq:theta_a}). Do other complex augmentations work for our model? To answer this question, we propose CBAM~\cite{woo2018cbam} (block attention) and FILM ~\cite{perez2018film} (linear transformation) weight augmentations inspired by TACT~\cite{choi2020visual}, which is a tracking method using different feature augmentations for context embedding. Specifically, assume the dimension of original weights $\theta_1$ is $m\times n$, \textit{i.e.}, $\theta_1\in \mathbb{R}^{m\times n}$, we can generate two attention maps $\delta_m\in \mathbb{R}^{m\times 1}$ and $\delta_n\in \mathbb{R}^{1\times n}$ by adjusting the output of prediction subnetwork $g_{\Delta}$. Then we obtain the adjusted weight by using the CBAM augmentation as follows $\theta_a=(\theta_1 \otimes \delta_m)\otimes \delta_n$, where $\otimes$ represents element-wise multiplication. Similarly, for FILM augmentation, we generate two coefficient maps $\gamma,\beta\in \mathbb{R}^{m\times n}$. Then we can obtain $\theta_a=(\theta_1 \otimes \gamma)+ \beta$.
}

To focus on the impacts of different augmentations on original CLNet, we omit the CU strategy and denote the tracker based on SiamRPN++~\cite{li2019siamrpn} as CLNet-RPN. We retrain CLNets by replacing additive augmentation with CBAM and FILM, and denote them as CLNet/CBAM and CLNet/FILM. Then we evaluate these models on the combined dataset~\cite{li2017visual,kiani_galoogahi2017need}. As shown in Table~\ref{tab:weight augmentation}, both CLNet/CBAM and CLNet/FILM obtain promising improvements than base tracker SiamRPN++ in terms of P and AUC, which indicates that our CLNet can utilize different weight augmentations. CLNet/FILM achieves better performance than CLNet/CBAM, since it uses more parameters for adjustment. However, CLNet-RPN with simple additive augmentation can still outperform them. The reason may be that CBAM and FILM are designed for feature augmentation, and they are not suitable for weight augmentation.
\begin{table}[]
\centering
		\caption{\small
		\textbf{Comparison of weight augmentation in terms of AUC and precision (P) scores on combined NfS30~\cite{kiani_galoogahi2017need} and DTB70~\cite{li2017visual} dataset.}  The base tracker is SiamRPN++~\cite{li2019siamrpn}. CBAM~\cite{woo2018cbam} and FILM~\cite{perez2018film} represent a block attention and linear transformation weight augmentation methods, respectively.
		}
	\label{tab:weight augmentation}
\setlength{\tabcolsep}{1mm}{
\begin{tabular}{ccccc}
\toprule
    & SiamRPN++ & CLNet/CBAM & CLNet/FILM & CLNet-RPN \\
    \midrule
P ($\%$)   & 69.9 & 72.2 & 73.4 & 74.4 \\
AUC ($\%$) & 55.8 & 58.4 & 58.7 & 60.0 \\
\bottomrule
\end{tabular}
}
\end{table}

\begin{table}[!tb]
\centering
		\caption{\small
		\textbf{AUC Scores of recent trackers on the LaSOT dataset~\cite{fan2019lasot}.} 
		}
	\label{tab:trans_light}
	\resizebox{0.5\textwidth}{!}{
\setlength{\tabcolsep}{0.5mm}{
    \begin{tabular}{cccccccc}
    \toprule
        ~ & TransT & TransT-M & STARK & E.T.Track & LightTrack & FEAR-L & CLNet*-BAN\\
        ~& \cite{chen2021transformer} &\cite{chen2022high} &\cite{yan2021learning} & \cite{blatter2021efficient} & \cite{yan2021lighttrack} &\cite{borsuk2021fear} &\bf Ours\\ \midrule
        AUC ($\%$) & 64.9 & 65.4 & 67.1 & 59.1 & 55.5 & 57.9 & 52.6\\ \bottomrule
    \end{tabular}
    }
    }
\end{table}

\subsection{More Discussions}
{To follow up the recent hot topic of transformer based tracking, we compare three representative trackers: TransT~\cite{chen2021transformer}, TransT-M~\cite{chen2022high}, and  STARK~\cite{yan2021learning} on LaSOT. As shown in Table~\ref{tab:trans_light}, there are gaps between transformer based models and our ResNet-based model (CLNet*-BAN), since former ones have stronger backbone. Besides, we also show the results of several efficient trackers: E.T.Track~\cite{blatter2021efficient}, LightTrack~\cite{yan2021lighttrack}, and FEAR-L~\cite{borsuk2021fear}, which have real-time speeds on the edge-platforms. In future work, we will try to incorporate our CLNet into these recent models with more efficient backbones, to explore a more generalized and practical solution.}

\section{Conclusion}\label{sec:conclusion}
{We have conducted an in-depth analysis to investigate Siamese-based trackers' performance degradation via visualization and tracking examples, and found that {\it missing decisive samples} during training is the critical factor. To alleviate it, we present a compact latent network (CLNet) to adjust Siamese-based models. Our CLNet is able to efficiently capture sequence-specific information from the initial frame. To further improve the training of adjusted networks, a new diverse sample mining approach is designed to enhance the discrimination ability.
Extensive experiments on three representative trackers have clearly demonstrated the generalization ability and effectiveness of our CLNet.}


%
\bibliographystyle{IEEEtran}
\small




















%

%

\vfill
%
%
%

\end{document}